\documentclass[colorlinks,citecolor=mydarkblue,urlcolor=mydarkblue,linkcolor=mydarkblue]{article}

\usepackage[dvipsnames]{xcolor}
\usepackage{graphicx} 
\usepackage{subcaption}

\usepackage[ruled,linesnumbered,boxed,commentsnumbered]{algorithm2e}
\usepackage{algorithmic}
\usepackage{amsmath}

\usepackage[final]{acl}
\usepackage{microtype}
\usepackage{graphicx}

\usepackage{booktabs} %
\usepackage{multirow}
\usepackage{multicol}
\usepackage{soul}
\usepackage{enumitem}
\usepackage{wrapfig}
\usepackage{float}
\usepackage[T1]{fontenc}
\usepackage{xcolor}         %
\usepackage{tcolorbox}
\usepackage{array}

\definecolor{bblue}{HTML}{4F81BD}
\definecolor{rred}{HTML}{C0504D}
\definecolor{ggreen}{HTML}{9BBB59}
\definecolor{ppurple}{HTML}{9F4C7C}
\definecolor{darkGreen}{rgb}{0.2,0.5,0.2}
\definecolor{mydarkblue}{rgb}{0,0.08,0.45}

\usepackage{pgfplots}
\usetikzlibrary{pgfplots.groupplots}
\pgfplotsset{compat=1.3}
\usepackage{tikz}
\usetikzlibrary{patterns}
\usepackage{caption}
\definecolor{myYellow}{rgb}{0.9,0.9,1}
\sethlcolor{myYellow}

\usepackage{soul}
\usepackage{amsmath}
\usepackage[utf8]{inputenc} 
\usepackage[T1]{fontenc}    
\usepackage{hyperref}       
\usepackage{url}            
\usepackage{booktabs}       
\usepackage{amsfonts}       
\usepackage{nicefrac}       
\usepackage{microtype}      
\usepackage{xcolor}         
\usepackage{commenting} 
\usepackage{fancyvrb}

\newcommand{\eat}[1]{}
\newcommand{\liangeat}[1]{}
\newcommand{\ie}{\emph{i.e.,}\xspace}

\title{A Hopfieldian View-based Interpretation for \\Chain-of-Thought Reasoning}

\author{
Lijie Hu$^{1,2}$,
Liang Liu$^{1,3}$,
Shu Yang$^{1,2}$,
Xin Chen$^{1,4}$,
Hongru Xiao$^{1,5}$,\\
\textbf{Mengdi Li$^{1,6}$,
Pan Zhou$^{1,7}$,
Muhammad Asif Ali$^{1,2}$,
and Di Wang}$^{1,2}$ \\
$^1$Provable Responsible AI and Data Analytics (PRADA) Lab\\
$^2$KAUST \quad $^3$Soochow University \quad $^4$University of Southampton 
$^5$Tongji University \\ 
$^6$Universität Hamburg \quad $^7$Huazhong University of Science and Technology
}

\begin{document}
\maketitle
\begin{abstract}
Chain-of-Thought (CoT) holds a significant place in augmenting the reasoning performance for large language models (LLMs). While some studies focus on improving CoT accuracy through methods like retrieval enhancement, yet a rigorous explanation for why CoT achieves such success remains unclear. 
In this paper, we analyze CoT methods under two different settings by asking the following questions:
(1) For zero-shot CoT, why does prompting the model with \emph{``let's think step by step''} significantly impact its outputs? 
(2) For few-shot CoT, why does providing examples before questioning the model could substantially improve its reasoning ability? To answer these questions, we conduct a top-down explainable 
analysis from the Hopfieldian view and propose a Read-and-Control approach for controlling the accuracy of CoT. Through extensive experiments on seven datasets for three different tasks, we demonstrate that our framework can decipher the inner workings of CoT, provide reasoning error localization, and control to come up with the correct reasoning path.



\end{abstract}

\section{Introduction}
Large Language Models (LLMs) have demonstrated exceptional 
capabilities in adhering to the natural language instructions \citep{ouyang2022instructgpt,mishra2022cross, WeiBZGYLDDL22,jin2024impact,yang2024human,yang2024moral}, and various downstream tasks \citep{kocon2023chatgpt, zhu2023multilingual, zhang2024unifying,hu2023differentially,yang2024dialectical}. An increasing amount 
of efforts are dedicated to exploring how these models can 
be utilized to perform more complex tasks, such as commonsense 
and mathematical reasoning~\cite{rae2021scaling, lu2023survey, imani2023mathprompter}. For this, Chain-of-Thought (CoT) techniques~\cite{wei2022chain, kojima2022large,
0002WSLCNCZ23, zhou2023least, qiao2023cotsurvey} have emerged 
as a simple yet effective method for enhancing the 
performance of LLMs in tasks requiring logical thinking. 

Recently, there have been multiple endeavors focused on improving the accuracy of CoT, \emph{e.g.,} efficient prompt design~\cite{zhang2023autoCoT, shi2023multilingual, wang2023ps}, process optimization~\cite{0002WSLCNCZ23, Madaan2023selfrefine, Shinn2023reflexion}, extra engine usage~\cite{Schick2023toolformer,Lyu2023faithfulcot,Gao2023pal}, knowledge enhancement~\cite{liu2022Genknow, yang2022logicsolver, he2023rethinking,Zhao2023verify}, \emph{etc}. 
Likewise, there have been numerous research attempts to identify the key factors or elements that help CoT to augment the reasoning capabilities of LLMs~\cite{kojima2022large, wang2023towards, tang2023large, merrill2023expresssive}. 
However, the majority of these studies primarily rely on designing datasets encompassing different features and/or using additional tools and knowledge to enhance the reasoning abilities of LLMs. 
Although some works have attempted to research the faithfulness of CoT~\cite{Lyu2023faithfulcot,Lanham2023CoTwithRL}, these studies have not fundamentally explained the key factors underlying the success of CoT.


\eat{
\begin{figure*}[t]
    \includegraphics[width=0.85\textwidth]{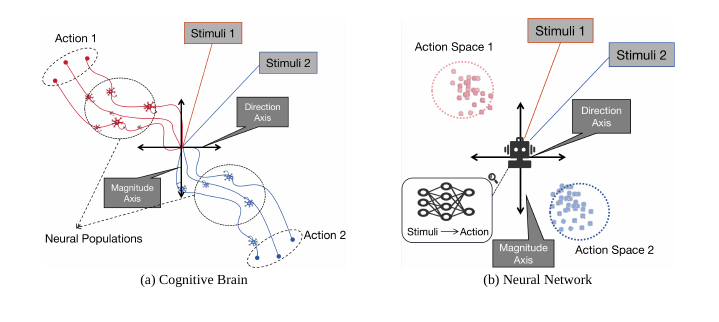} 
\end{figure*}
}

\begin{figure*}[t]
\centering
\begin{minipage}[t]{0.48\textwidth}
  \centering
  \includegraphics[width=0.80\textwidth]{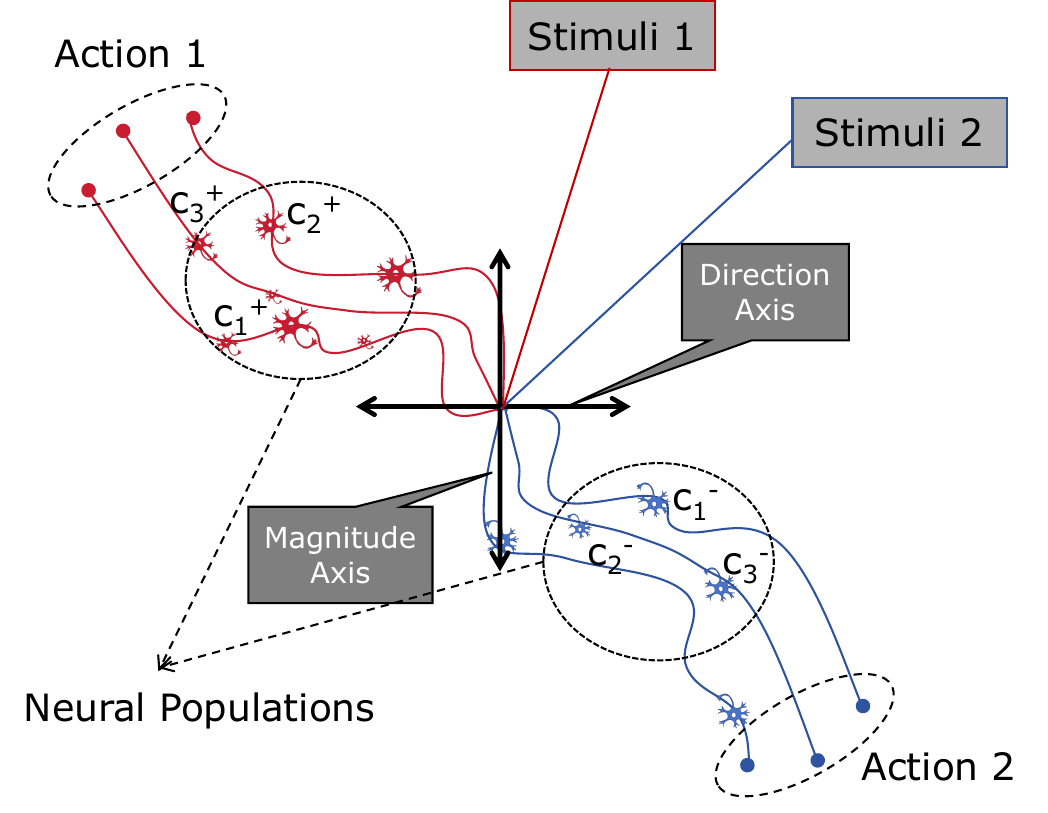} 
  \subcaption{Cognitive Brain}
  \label{subfig:cb}
\end{minipage}
\hfill
\begin{minipage}[t]{0.40\textwidth}
  \centering
  \includegraphics[width=0.80\textwidth]{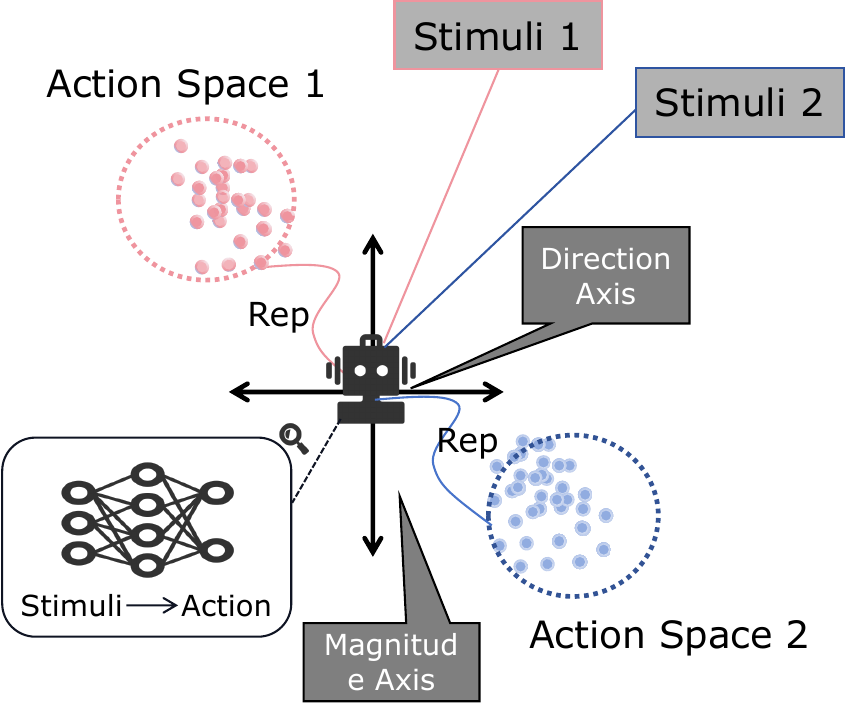}
  \subcaption{Neural Network}
  \label{subfig:nn}
\end{minipage}
\caption{Illustration of mechanisms in the cognitive brain~\cite{barack2021twoviews} and neural network.}
\label{fig:hopfieldianandstumili}
\vspace{-15pt}
\end{figure*}


To bridge this gap, in this work, we propose an explainable 
framework to identify the pivotal elements influencing CoT 
performance for LLMs. Specifically, we build our framework 
on the top of~\emph{Hopfieldian view}~\cite{barack2021twoviews}. 
As shown in Figure~\ref{subfig:cb}, the \emph{Hopfieldian view} 
explains cognition as the result of transformations between, 
or movements within, representational spaces implemented by 
\textit{neural populations} in response to \emph{stimuli}. 
In this view, we no longer consider the interaction process 
between neurons, rather we only need to understand the impact 
of the representational changes on the final actions, thus 
offering a natural setting to study LLMs.


As shown in Figure~\ref{fig:framework}, our framework 
encompasses three core components: 
(i) \emph{Concept Modeling},  
(ii) \emph{Concept Simulation}, and 
(iii) \emph{Analysis based on Hopfieldian View}.
The \emph{Concept Modeling}\eat{ (Section \ref{sec:stage1})} 
emphasizes that during pre-training
LLMs learn latent concepts (both concrete and abstract) that pertain to specific names or domains\eat{ and also represent more 
abstract notions derived from document-level modeling}. 
The \emph{Concept Simulation} \eat{(Section \ref{sec:stage2}),} 
highlights that we can use zero-shot or few-shot prompts as stimuli 
for the LLMs in order to induce and/or activate specific concepts. 
These activated concepts collectively determine the final action 
of the model.
Finally, we perform analysis based on \emph{Hopfieldian view} 
to analyze the representations responsible for  the activation of 
concepts when the model is stimulated.
Specifically, it uses a \underline{\emph{Read}} operation to read 
concept-specific representations for reasoning and error
localization; followed by a \underline{\emph{Control}}
operation to guide and/or rectify the reasoning direction of LLM.

We apply our proposed approach to answer the following 
key questions\eat{ regarding the interpretability of CoTs}: 
(i) For zero-shot CoT~\cite{kojima2022large}, why prompting 
the model with stimulus like \emph{``let's think step by step''} before inference can make LLMs' outputs more accurate? (ii) For few-shot CoT~\cite{wei2022chain}, why providing example demonstrations before questioning significantly 
enhances the LLMs' reasoning ability? 

Experimental evaluation on three different tasks: \emph{i.e.,} (i) arithmetic reasoning, (ii) commonsense reasoning, and (iii) symbolic reasoning, reveals that our framework can provide intuitive and interpretable analysis for CoT reasoning helpful in tracing and control the errors LLMs made during the CoT process. 

We summarize the key contributions of our work as follows:
\begin{enumerate}
\itemsep0em
    \item We propose a novel framework employing \emph{Hopfieldian view}
    to identify the key factors responsible for the success of CoT under 
    zero-shot and few-shot settings. 
    \item To the best of our knowledge, this work is an initial attempt at interpretability in CoT 
    that employs concept-level representation read operations to 
    localize errors in CoT followed by control operation to rectify 
    LLMs' reasoning paths. 
    \item We conduct a comprehensive experimental evaluation for
    our framework by employing seven different data sets across three different tasks. Results showcase that our framework can offer faithful explanations with error localization and control for CoT. 
\end{enumerate}

\section{Related Work}
\subsection{Chain of Thought (CoT)}

The CoT is a prompting technique that 
engages LLMs in step-by-step reasoning rather than directly providing the answers~\cite{nye2021work}. Studies show that intermediate steps or learning from demonstration can significantly improve the reasoning performance of LLMs~\cite{wei2022chain,kojima2022large}. 
Owing to the success of CoT, numerous studies focus on using CoT to solve complex problems, such as commonsense, arithmetic, symbolic reasoning~\cite{0002WSLCNCZ23, zhou2023least, Wang2024noprompting}, and logic tasks~\cite{Creswell2022faithfulforlogic, Pan2023logiclm, weng2023selfv}.\eat{Also, there have been attempts to extend the core concepts of CoT to multimodal settings~\cite{zhang2023mcot, zheng2023mcot, zhang2023mcotKG}.}
Recently, numerous endeavors have been made to identify the key factors through which CoT enhances the reasoning capabilities of LLMs~\cite{wang2023towards, Dutta2024howthink}. 
For instance,~\citet{kim-etal-2023-cotever} corrected the erroneous parts of the chain by using a query-based approach.
\citet{Zhao2023verify} proposed a knowledge-enhanced method to augment the factual correctness for multi-pole open-domain QA tasks. 
\citet{Lyu2023faithfulcot} introduce faithful CoT, ~\emph{i.e.,} a faithful by-construction framework to first translate natural language query to symbolic reasoning chain, later solve the problem via with CoT.

Likewise, numerous research attempts have focused on the sequence and quantity of demonstrations within the context, investigating their contributions to the final reasoning performance.
For this,~\citet{min2022rethinking} discovered that even random labels or ineffective reasoning steps can still improve the model's reasoning performance.
\citet{Lanham2023CoTwithRL} demonstrated the impact 
of intervening in the CoT process by adding mistakes or paraphrases.
\citet{pfau2024letdot} showed that using meaningless filler tokens in place of a chain-of-thought can help improve the performance.

These studies have led to concerns such as whether LLMs truly acquire the ability to learn inductive, deductive, and abductive reasoning skills from demonstrations and instructions~\cite{stechly2024chain,jin2024impact}. In this paper,  we will use a top-down interpretability method to analyze the operating mechanism of CoT reasoning.

\subsection{Interpretability of LLMs}
Interpretability plays a key role in a deeper understanding 
of LLMs to identify potential risks and better meet human requirements~\cite{zou2023representation}. Current widely used strategies for interpretability include: 
(i) salience maps, relying on highlighting the regions in the input that are attended by the model
~\cite{2013_deep, 2017_smoothgrad, 2019_does,hu2023seat,hu2023improving,lai2023faithful}; 
(ii) feature visualization, creating representative inputs indicative of particular neurons' activations~\cite{2013_intriguing, 2016_synthesizing, 2019_understanding,
2018_net2vec}, and
(iii) mechanistic interpretability, employing reverse-engineering
tools to explain the network in terms of circuits and node-to-node connections~\cite{olah2020zoom, olsson2022context, wang2022interpretability}. 
However, these methods require a lot of human intervention and 
are limited in terms of interpretability, especially for the 
neural network models~\cite{jain-wallace-2019-attention,2018_net2vec,hu2024editable}. Thus, these methods cannot be directly used to interpret CoTs.
Meanwhile, current approaches are limited to the analysis at the representation level and do not consider the relationship between representations and the features/concepts learned during pre-training~\cite{Bricken2023Monosemanticity,Templeton2023Scaling}.

Other dominant works in this regard investigate the location and representation of concepts in the 
network~\cite{2018_interpretability,li2024text}, linear classifier probing to probe properties in the input~\cite{2022probing}, locating and editing facts~\cite{2022_locating, 2023_mquake,cheng2024leveraging,cheng2024multi}, concept erasure~\cite{2022_gold, 2023_erasing}, corrective 
analysis~\cite{Burns2022DiscoveringLK}, \textit{etc.} These observations are aligned with RepE~\cite{zou2023representation} that emphasized that representations within the LLM are almost secretly linear~\cite{Park2023linearhp}. However, none of them consider the inner workings of the Chain-of-Thought Reasoning.


\section{Preliminaries}
\label{sec:prelimnaries}

In this section, we briefly introduce the notation 
followed by a quick background on the core 
concepts required for this work. More background can be found in Appendix \ref{cot}. 


\noindent{\bf Notations.} We use $\mathcal{M}_{\theta}$ to 
denote an LLM parameterized by $\theta$,  
$C$ to represent set of concepts, and $c_i$ an individual concept, 
$T$ to represent the prompt template, $\mathcal{S}$ to denote 
a set of \emph{stimulus} $s$ in the prompt (\emph{i.e.,} 
example input-output demonstrations in a few-shot scenario 
and \emph{``think step by step''} in a zero-shot scenario), 
and $x$ denotes the query for which the LLM is asked to provide 
a response. Thus, we can use $T([\mathcal{S}, x])$ to denote 
the prompt $p$, which includes both a set of stimuli 
$\mathcal{S}$ and query $x$, providing the full context to 
LLM in order to generate a response.

\liangeat{
\warn{
$C$ to represent set of concepts, and $c_i$ an individual concept.
$s \in S$ represents stimulus, \emph{i.e.,} that can be an action, or prompt etcx...!
}
}

\paragraph{Hopfieldian View.}

Hopfieldian view aims to scale/extend the traditional 
mechanistic interpretability for AI-cognition 
(\emph{i.e.,} neurons and circuits) to a much broader perspective in attempts to explain complex phenomena.  
It considers representation as a basic unit and explains cognition as the result of transformations between or movement within representational spaces implemented via neural populations~\cite{barack2021twoviews}. 

\eat{Figure~\ref{subfig:nn} indicates that in a large model, 
each action can be seen as a neural network responding to a 
stimulus (in our work, a key part of prompts see Section~\ref{sec:method} 
for details) and arriving at a specific \emph{action space}. 
Instead of analyzing specific sub-networks or algorithms within the model, we need to understand the types of \emph{representational changes} that guide the model's final output into different action spaces.}

We provide an intuitive explanation in this regard in Figure~\ref{fig:hopfieldianandstumili}, which 
emphasizes that Hopfieldian view allows us with the provision to analyze and control the network as a 
function of external stimulus the akin to the 
functionality of cognitive brain.
For further details refer to recent work by \citet{zou2023representation} that employed Hopfieldian view to propose a top-down approach to interpretability.

\begin{figure*}[t]
    \centering
    \vspace{-3.7ex}    \includegraphics[width=0.79\linewidth]{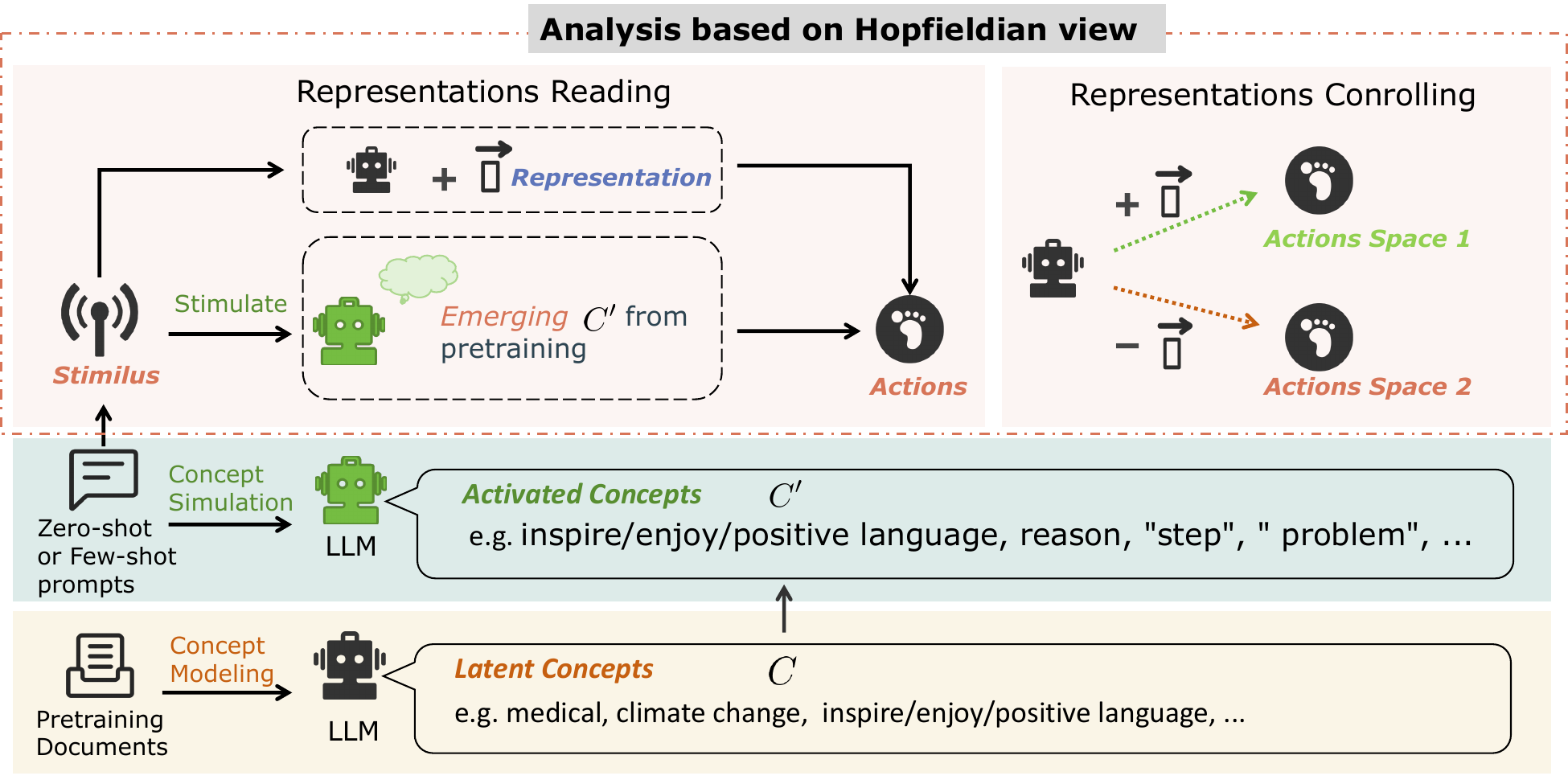}
    \caption{Our CoT explanation framework based on Hopfieldian view.
    \liangeat{
    LLM learned latent (both concrete and abstract) concepts from pertaining documents. These concepts might pertain to specific names or domains, such as ``medical'' or ``climate change''. They could also represent more abstract notions derived from document-level modeling, such as ``positive language'', ``careful reasoning'', or ``affirmative expression''. During the reasoning process, zero-shot or few-shot prompts serve as stimuli for the LLM, inducing the activation of specific concepts. These activated concepts collectively determine the final action of this model. However, directly extracting activated concepts would consume a tremendous amount of computational resources, and individual concepts alone cannot determine the model's overall action. Therefore, we use representations to denote the activation of concepts when the model is stimulated.}}
    \vspace{-3.7ex} 
    \label{fig:framework}
\end{figure*}

\paragraph{In-context Learning(ICL) and Bayesian Inference.}
Bayesian inference is a method for statistical inference that is used to compute the posterior probability based on the likelihood and prior. It has gained attraction in the analysis and interpretation of LLMs in order to systematically update the knowledge and beliefs for a network with the acquisition of new information. Further details exhibiting the impact of Bayesian analysis peculiarly focued on LLMs is available in~\cite{LLM_Bayesian}. 

The in-context learning ability of models may also 
be represented via Bayesian inference as follows:
\begin{equation}
\label{eq:bayesian_1}
\vspace{-0.1ex}
\mathbb{P}(r | p) =  \int_{c} \mathbb{P} (r |c, p)  \mathbb{P} (c | p) \mathop{d}(c).
\vspace{-0.1ex}
\end{equation}
This equation illustrates the \emph{conditional probability} of generating a response $r$ given a prompt $p$, as a function of the concepts $c$ that the model has learned during pre-training. By integrating over these concepts, the model dynamically adjusts its responses based on the likelihood of various concepts being relevant to the given prompt. This probabilistic approach not only helps in understanding the model's decision-making process but also in enhancing its performance on complex reasoning tasks. Our work extends this formula by incorporating the Hopfieldian view, describing the role of prompts in model reasoning as stimuli that activate specific populations of neurons.

\section{CoT Explanation from Hopfieldian View}
\label{sec:method}

In this section, we will dive into details about our motivation from the  Hopfieldian view and a general introduction of our framework.


\noindent{\bf Motivation.}
Existing literature has confirmed that CoT can improve the reasoning ability of LLMs~\cite{kojima2022large,wei2022chain}. 
However, few studies have conducted a comprehensive analysis of the interpretability of CoT.  This is mainly because reasoning is a complex task, and the model's final performance is influenced by numerous factors. Analyzing internal activations in isolation or from the perspective of individual neurons of LLMs is challenging. Inspired by work related to representation engineering~\cite{zou2023representation}, we consider LLM from a higher-level cognition perspective (Hopfieldian view), \emph{i.e}., specific external stimuli activate representations within the model, thereby causing the model to exhibit different behaviors that represent the population activation behavior of neurons within the model.
Meanwhile, many studies have shown that the representations of LLM are almost secretly linear inside, and can be controlled via linear probe methods~\cite{Park2023linearhp,wu2024reft}.
These phenomena indicate that we can analyze the interpretability of CoT from the perspective of representation and control the reasoning process of LLM.

\liangeat{
LLM learned latent (both concrete and abstract) concepts from pertaining documents. 
These concepts might pertain to specific names or domains, such as ``medical'' or ``climate change''. 
They could also represent more abstract notions derived from document-level modeling, such as ``positive language'', ``careful reasoning'', or ``affirmative expression''. 
During the reasoning process, zero-shot or few-shot prompts serve as stimuli for the LLM, inducing the activation of specific concepts. 
These activated concepts collectively determine the final action of this model.
However, directly extracting activated concepts would consume a tremendous amount of computational resources, and individual concepts alone cannot determine the model's overall action. 
Therefore, we use representations to denote the activation of concepts when the model is stimulated.
}

\noindent{\bf Framework.}
As we are targeting at exploiting the Hopfieldian view to provide an explanation for LLMs' CoT reasoning, we have to build a bridge from the components in the cognitive brain to CoTs. We begin from introducing the definition of stimulus in CoT, which is the prompt text (\emph{i.e.,} example input-output demonstrations in a few-shot scenario and \emph{``think step by step''} in a zero-shot scenario). Then use these stimuli to activate specific "neural populations" for CoT.

For neural populations, ideally, we can leverage humanly comprehensible concepts within LLMs' representations. To model these concepts, our framework has two phases: 
(i) \emph{Concept Modeling},  and (ii)\emph{Concept Simulation}. 
In the Concept Modeling phase (Section \ref{sec:stage1}), LLMs will learn latent (both concrete and abstract) concepts from pertaining documents, and these concepts might pertain to specific names or domains, such as ``medical'' or ``climate change''. 
They could also represent more abstract notions derived from document-level modeling, such as ``positive language'', ``careful reasoning'', or ``affirmative expression''. 
Likewise, in the Concept Simulation stage (Section \ref{sec:stage2}), zero-shot or few-shot prompts serve as stimuli for the LLM, inducing the activation of specific concepts. 
These activated concepts collectively determine the final action of this model.

However, directly extracting activated concepts would consume a tremendous amount of computational resources, and individual concepts alone cannot determine the model's overall action. Therefore, our framework uses representations to denote the activation of concepts when the model is stimulated. For this, our framework employs \emph{Representations  Reading} (Section \ref{sec:stage3}) and \emph{Representations Controlling} (Section \ref{sec:stage4}) as the core modeling components.  

Next, we will give details about individual model components. Later, we explain how this framework can be used to improve the CoT reasoning abilities of LLMs. See Figure \ref{fig:framework} for an illustration of our framework.  

\section{Details of the Framework}

\subsection{Concept Modeling}\label{sec:stage1}
Previous research~\cite{xie2021explanation} reveals that during pre-training LLMs can learn a set of latent document-level concepts $c \in C$ that in turn help large language models to generate coherent next tokens. This motivates our design of stimuli. 
We provide some examples of these concepts in Figure~\ref{fig:framework}.
The activation and intensity of specific concepts determine the model's final response. 
However, searching and controlling individual 
concepts is a prohibitively expensive process. 
In order to circumvent that, we instead use representations 
as a basic unit indicative of latent concepts 
learned by the model. We cover the definitions and relationships between these representations and concepts in the next subsections.




\subsection{Concept Stimulation}\label{sec:stage2}

From the Hopfield view, we need specific stimulus to trigger the model's specific action. Since CoT reasoning is primarily activated through prompts, we posit that CoT prompts $p$ containing a set of stimulus $S$ that can activate the model's reasoning capabilities. So, we use the following formula based on Bayesian inference to illustrate the entire process from the LLM receiving an input to generating a specified output, as shown in Figure~\ref{fig:framework}:

\begin{equation}
\label{eq:bayes_2}
\mathbb{P}(r | p) =  \int_{c} \mathbb{P} (r |c,T([\mathcal{S}, x]))  \mathbb{P} (c | T([\mathcal{S}, x])) \mathop{d}(c),
\end{equation}
where, $T([S, x])$ is the prompt template based on
specific Stimulus $S$ and query $x$. 
$\mathbb{P} (r |c,T([\mathcal{S}, x]))$ is the 
likelihood of response $r$ given $c$, and 
$\mathbb{P} (c | T([\mathcal{S}, x]))$ is the prior
probability of the $c$ for a given prompt template
$T([\mathcal{S}, x])$.

This equation governs the conditional probability of a response $r$ given prompt $p$ guided by concept 
\emph{i.e.,} $(c \in C)$ learned by the model. 



\subsection{Representations Reading}\label{sec:stage3}
The end goal of reading representations is to look
for and/or locate the representations for key 
concepts within a network.
We sub-divided its process into: 
(i) concepts reading, and 
(ii) reasoning error localization. 
Details about each component are explained in the following sub-sections.

\subsubsection{Concepts Reading}
\label{sec:reading}
This module aims to locate and highlight the core 
concepts within the network. The concepts output/generated by this module act as a foundation for the subsequent steps, \ie representation evaluation 
followed by representation control.
However, in CoT-style prompting, the definition of concepts 
is abstract, and how to define high-level concepts is crucial.
For this, we utilize a Linear Artificial Tomography (LAT), 
similar to \cite{zou2023representation}, in order to 
identify the directions of the key concept, as illustrated 
in Algorithm~\ref{alg:LAT}.
It encompasses two different modules:
(i) Stimuli Selection; 
(ii) Neural Activity Monitoring; and
(iii) Concept Identification, with each module 
explained as follows:

\begin{algorithm}[t]
    \caption{LAT Process-flow}
    \label{alg:LAT}
    \KwIn{Stimulus set $S$, model $\mathcal{M}_{\theta}$, function $Rep(.,.)$}
    \KwOut{Reading vector $v$}

    $A_c \gets []$ \tcp{neural activity list}
    
    \For{$s_i$ \textbf{in} $S$}{
        $A_c \gets A_c + [Rep(s_i, \mathcal{M}_{\theta})[-1]]$ 
    }
    
    $i \gets 0$
    
    $D \gets []$ \tcp{difference list}
    
    \While{$i < len(S) - 1$}{
        $D \gets D + [A_c[i+1] - A_c[i]]$ \label{line:diff}

        $i \gets i + 2$
    }

    $v \gets \text{PCA}(D)[0]$

    \textbf{return} $v$
\end{algorithm}













\begin{algorithm}[b]
   \caption{Control Representations}
    \label{alg:control}
    \KwIn{Query $x$, 
    reading vector $v$, 
    control layers $L$, 
    model $\mathcal{M}_{\theta}$}
    \KwOut{Model's logits after control}

    \tcp{get the representations of $x$}
    $H \gets Rep(x,\mathcal{M}_{\theta})$ 

    \For{$layer$ \textbf{in} $\mathcal{M}_{\theta}$}{
    \If{$layer$ \textbf{in} $L$}{
            $H[layer] \gets H[layer] + v[layer]$ \label{line:addv}
        }
    }
    \tcp{transform $H$ into the logits of model}
    $logits \gets \text{trans\_into\_logits}(H)$
    
    \textbf{return} $logits$
\end{algorithm}





\begin{figure*}[t]
    \centering
    \includegraphics[width=0.80\linewidth]{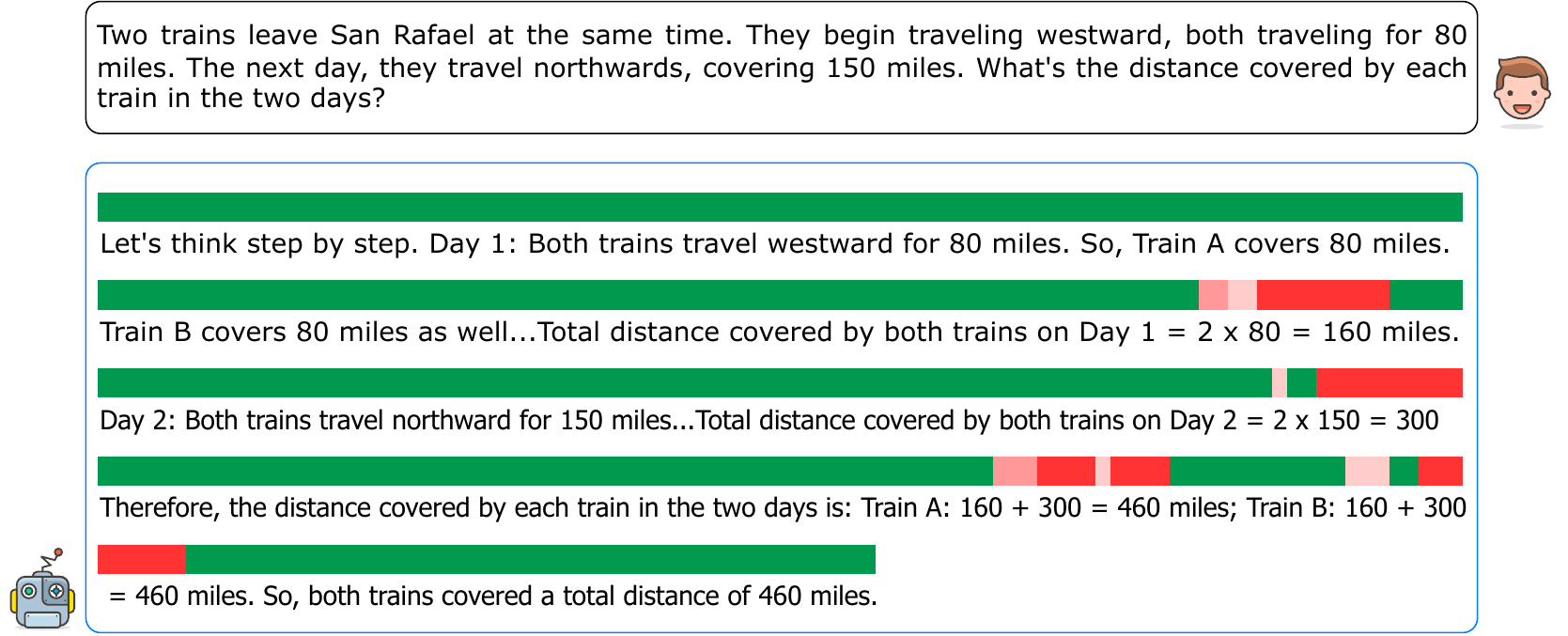}
    \caption{A real case of reasoning error localization by using LLaMA-2-7B-chat in a zero-shot scenario for the GSM8K dataset using our framework. The \textcolor{OliveGreen}{\textbf{green}} bar indicates that the reasoning snippet is correct, and the \textcolor{Red}{\textbf{red}} bar means that the reasoning snippet may be wrong.   }
    \label{fig:reading_case}
\vspace{-7pt}
\end{figure*}

\paragraph{(i) Stimuli Selection.}
\label{sec:select_stimuli}

To elucidate the sensitivity of LLMs to different CoT prompts, in this step we will employ various ``\emph{stimuli}'' encompassing one or more concepts for inducing diverse model behaviors under both zero-shot and few-shot settings. Our objective is to analyze how the concepts corresponding to a specific stimulus are processed and transferred within large models, as illustrated in Figure~\ref{fig:hopfieldianandstumili}, from the perspective of cognitive neuroscience.

We implement the impact of stimulus in Hopfieldian 
view as follows: a set of stimuli (i.e., the prompt text) $S = \{s_{0}^{-}, s_{0}^{+}, s_{1}^{-}, s_{1}^{+}, \cdots, s_{n}^{+}\}$, when an LLM receives a positive stimulus 
$s^+$, it generates a specific output/response $r$, across 
a query $x$, and falls into the corresponding action 
space $A^+$. 
Conversely, when it receives a negative stimulus $s^-$ or no stimulus at all, the model's output falls into a different action space $A^-$ or $A$.

\liangeat{
For this, we aim to analyze, why providing 
examples or a prompt text (\ie ``Let's think step by step'') can improve the reasoning accuracy of LLMs.
}
An example of a pair of specific prompt template for $T([\mathcal{S}, x]))$, and example demonstrations are shown in Appendix~\ref{Appendix:prompts}.

\paragraph{(ii) Neural Activity Monitoring.}
After receiving the task-specific prompt stimuli set, 
the next step is to capture corresponding representations 
from the LLM. Given that LLMs
rely on transformer-based architecture to store distinct representations intended for different purposes. 
In order to capture the representations for the specific 
task, we first need to identify suitable design choices for the 
extraction process.
For decoder-only/auto-regressive architecture models, one 
natural choice is to choose the position of the \emph{last token} as the representation of each concept. 

Formally, given a set of stimuli $S$, a language model $\mathcal{M}_{\theta}$, and a function $\text{Rep}$,  we compute the neural activity $A_c$ as follows:
\begin{equation}
\label{Eq:NA}
    A_c = \{\text{Rep}(s_i, \mathcal{M}_\theta)[-1] \;|\; s_i \in S \},
\end{equation}
where $\text{Rep(.,.)}$ is a function that takes stimulus $s_i \in S$ and LLM $\mathcal{M}_{\theta}$ as input and returns the representations from all token positions. 

\eat{
\begin{equation}
    A_c = \text{mean}\{(\text{Rep}(s_i, \mathcal{M}_\theta)[\text{LOC}_c]) \;|\; s_i \in T_c \}
\end{equation}
}
\paragraph{(iii) Concept Identification.}\label{subsec:vector}
Concept identification aims to identify a direction that accurately predicts the underlying concept solely based on the neural activity recorded via~\eqref{Eq:NA}.

\liangeat{
For this, we assume that the concepts may be represented 
as approximately linear representations \di{why you can assume this, is this reasnonable?}, 
}

\liangeat{For this, we use Principal Component Analysis (PCA) to model them \di{model what?}.}

For this, we employ Principal Component Analysis (PCA)  to the set of difference vectors that can yield superior results when the stimuli in each pair share similarities except for the target concept.
Specifically, we compute the principal components of the representation $\{ A_{c}^{i+1} - A_{c}^{i} \}$ (as depicted in line~\ref{line:diff} of Algorithm~\ref{alg:LAT}),
and define the leading principal vector derived 
from this as ``reading vector'', denoted by $v$.
Intuitively, we can compute the dot product between the 
reading vector $v$ and each representation $\text{Rep}(s_i, \mathcal{M}_\theta)$ as an indicator of the relevance 
between the stimulus representation and latent concepts.


\subsubsection{Reasoning Error Localization}
\label{sec:EV}


In this section, we consider the visualization of reasoning error localization, 
given a query $x$, we format it using the stimulus prompting template $T$ and represent it as $p$. Then, we use $p$ to generate a base reasoning path $R$ with $\mathcal{M}_\theta$.

After the representations reading (section~\ref{sec:reading}), we get the reading vector $v$.
Then, we compute the dot product between the representations $\text{Rep}(p,\mathcal{M}_\theta)$ and our reading vector $v$ yielding
a set of scores, which serve as the basis for reasoning error localization. We use the score of each token compared against a threshold $\delta$ to locate errors in the reasoning path.
Specifically, we use the following criterion to access and/or evaluate the quality of the rationales:
\begin{equation}
    \text{scores}_{norm} = 
    \text{normalize}(\text{Rep}(R, \mathcal{M}_{\theta})^T v - \delta).
\end{equation}

Note, in the above equation, using multiple different values 
for $\delta$ provides us with the provision to effectively 
adjust and use a wide range of scores.
Then, we use a clip function to set the value greater 
than 0 in the score array, \emph{i.e.,} $(\text{scores}_{norm})$ to 0. 
Our salience map is based on this score, therefore, when the score is less than 0, it will appear in red, indicating that there is a potential wrong reasoning location.

An example of visualization is shown in 
Figure~\ref{fig:reading_case}, LLaMA-2-7B-chat did not really understand the meaning 
of the given question and misinterpreted 
``\emph{the distance covered by each train}'' as 
``\emph{the total distance covered by both trains}'', 
thus giving an incorrect reasoning path.

\subsection{Representations Controlling}\label{sec:stage4}
\label{sec:RC}

Here we aim to correct the direction of the concept which may lead to CoT errors, as shown in the top right part of Figure \ref{fig:framework}. 

Given a query $x$, we adopt a simple process to control the direction of CoT. 
Specifically, as depicted in Algorithm~\ref{alg:control}, during the generation phase, we first acquire the reading vector $v$ from the concept identification phase.
Subsequently, we get the representations $H$ for the query $x$, and then we incorporate $v$ into the original representation weights of the corresponding control layers $L$ (as shown in line~\ref{line:addv} of Algorithm~\ref{alg:control}).
Finally, we transform the controlled representations $H$  into the logits of model for generation.

\begin{figure*}[ht]
    \centering
    \includegraphics[width=0.9\linewidth]{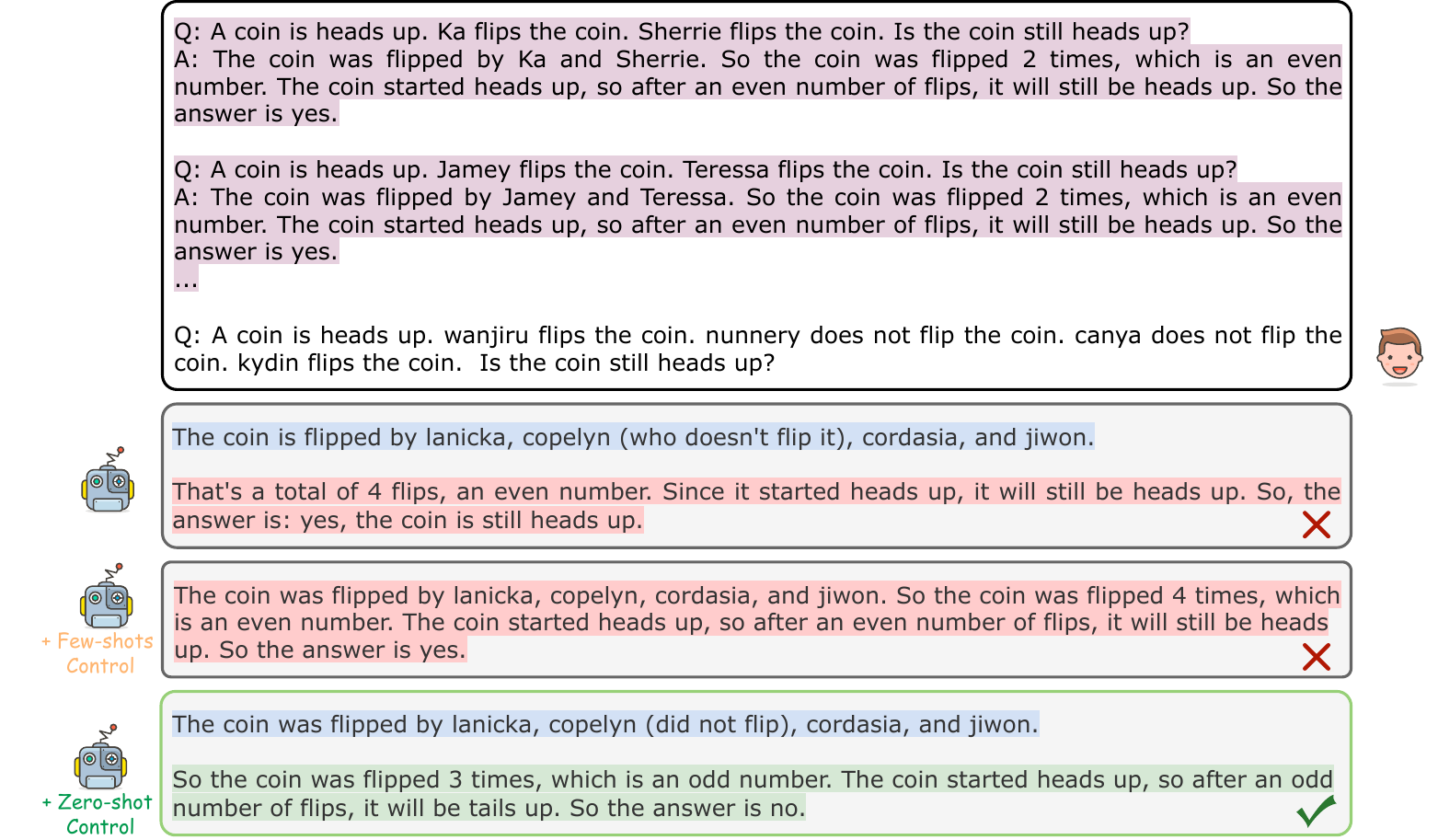}
    \caption{A real case predicted by LLaMA-3-8B-instruct with few-shot CoT on the coin flip dataset. 
    The \textcolor{purple}{purple} part is an example of input-output pairs given by user.
    The segment highlighted in \textcolor{skyblue}{blue} represents the correct output of the model. 
    The \textcolor{crimson}{red} part shows that the model starts to reason in the wrong direction without control, 
    while the \textcolor{seagreen}{green} portion indicates the model reason in the correct direction after adding control.}
    \label{fig:zero_few_control}
\vspace{-12pt}
\end{figure*}

\section{Experiments}
\subsection{Experimental Setup}
\noindent{\bf Dataset.}
We consider 7 datasets for 3 different tasks: Arithmetic Reasoning, Commonsense Reasoning, and Symbolic Reasoning. Specifically, for arithmetic reasoning, we select the GSM8K~\cite{cobbe2021gsm8k}, SVAMP~\cite{patel-etal-2021-nlp} and AQuA~\cite{miao-etal-2020-diverse} datasets; for commonsense reasoning, we choose the benchmarks of strategyQA~\cite{10.1162/tacl_a_00370} and CSQA~\cite{talmor2019commonsenseqa}; and for symbolic reasoning, we pick the Coin Flip~\cite{wei2022chain} and Random Letter datasets (constructed from the last letter dataset~\cite{wei2022chain}). Further details and statistics of the data sets are given in 
Appendix~\ref{Appendix:data}. 

\eat{\noindent{\bf Settings.}
Prompt. We analyze\liangeat{ed} the performance of the models using zero-shot and few-shot prompts.
\warn{For the zero-shot scenario, we utilize the most common 
zero-shot CoT~\cite{kojima2022large}, to provide the thought 
path by simply adding (\emph{Let's think step by step}) before answering. For few-shot, we feed in the $n$ different examples prior to generating a response from LLM.
Example illustrations in this regard are provided in the 
Appendix~\ref{Appendix:Prompts_E}.}}

\liangeat{
For zero-shot prompts, we also added the text \emph{``Let's think step by step''} 
before the question in order to strengthen the model's step-by-step 
reasoning abilities. For few-shot prompts, we feed in the quiz in a 
sequence and use chain-of-thought reasoning to generate the answer. 
Example illustrations in this regard are provided in the 
Appendix~\ref{tab:appendix-coinflip-prompt}.}

\noindent{\bf Baselines.}
We use three LLMs for evaluation, \ie 
(i) Mistral-instruct-7B~\cite{jiang2023mistral},
(ii) LLamA-2-7B-chat~\cite{touvron2023llama},
(iii) LLamA-3-8B-instruct~\cite{metallama3}.
\liangeat{, and (iv) CodeQwen1.5-7B~\cite{bai2023qwen}.}

\eat{We use four LLMs for evaluation, \ie 
(i) Mistral-instruct-7B\footnote{https://huggingface.co/mistralai/Mistral-7B-Instruct-v0.2}~\cite{jiang2023mistral},
(ii) Llama2-7B-chat\footnote{https://huggingface.co/meta-llama/Llama-2-7b-chat-hf}~\cite{touvron2023llama},
(iii) Llama3-8B-instruct\footnote{https://huggingface.co/meta-llama/Meta-Llama-3-8B-Instruct}~\cite{metallama3},\fixshu{Add citation}
and 
(iv) CodeQwen1.5-7B\footnote{https://huggingface.co/Qwen/CodeQwen1.5-7B-Chat}~\cite{bai2023qwen}.}

\noindent{\bf Evaluation Metrics.}
We use accuracy metrics for all dataset benchmarks. The answer extraction process is based on the methodology outlined by~\citet{kojima2022large}. Detailed procedures and results are provided in the Appendix~\ref{Appendix:extract}.

\noindent{\bf Implementation Details.}
In our study, we pick the last ten layers to control the reasoning direction of the LLM and select different numbers of stimulus samples according to different tasks. Specifically, We conducted experiments using three distinct reading set sizes: 128, 256, and 512 samples, details shown in Appendix~\ref{Appendix:reading_size}. 
For experimentation, we use a value of threshold $\delta$ = 3.5.
We use float16 to load large models. We employ greedy search as our decoding strategy and set max new tokens of 512 for all datasets. All experiments are conducted 
using NVIDIA L20 GPU.

\begin{table*}[ht]
\centering
\resizebox{0.8\textwidth}{!}{
\begin{tabular}{lcccc}
\hline
\multirow{2}{*}{Model} & \multicolumn{3}{c}{Arithmetic Reasoning} \\ \cline{2-4} 
 & GSM8K & SVAMP & AQuA \\ \hline
\multicolumn{4}{c}{zero-shot CoT} \\ \hline
Mistral-7B-instruct & 45.03 / \textbf{47.08} & 57.33 / \textbf{61.33} & 25.59 / \textbf{30.71} \\
LLaMA-2-7B-chat & 26.31 / \textbf{27.37} & 46.00 / \textbf{46.67} & 27.95 / \textbf{30.71} \\
LLaMA-3-8B-instruct & 74.45 / \textbf{75.89} & 82.67 / \textbf{84.00} & 42.13 / \textbf{45.67} \\ 
Avg. & 48.60 / 50.11 (\textbf{1.51↑}) & 62.00 / 64.00 (\textbf{2.00↑}) & 31.89 / 35.70 (\textbf{3.81↑})  \\ \hline
\multicolumn{4}{c}{few-shot CoT} \\ \hline
Mistral-7B-instruct & 46.78 / \textbf{47.16} & 61.33 / \textbf{61.67} & 34.65 / \textbf{35.04} \\
LLaMA-2-7B-chat & 5.00 / \textbf{5.16} & 38.67 / \textbf{39.00} & 22.44 / \textbf{23.62} \\
LLaMA-3-8B-instruct & 72.71 / \textbf{74.83} & 82.00 / \textbf{83.67} & 52.76 / \textbf{53.21} \\
Avg. & 41.50 / 42.38 (\textbf{0.88↑}) &  60.67 / 61.45 (\textbf{0.78↑}) & 36.61 / 37.29 (\textbf{0.68↑}) \\ \hline
\end{tabular}
}
\caption{Results on the Arithmetic Reasoning task for different models. The left side shows scores without control, and the right side shows scores after control.}
\label{tab:arithmetic_result}
\vspace{-7pt}
\end{table*}
\begin{table}[ht]
\centering
\resizebox{\columnwidth}{!}{
\begin{tabular}{lcc}
\hline
\multirow{2}{*}{Model} & \multicolumn{2}{c}{Commonsense Reasoning} \\ \cline{2-3} 
 & StrategyQA & CSQA \\ \hline
\multicolumn{3}{c}{zero-shot CoT} \\ \hline
Mistral-7B-instruct & 63.01 / \textbf{63.97} & 67.57 / \textbf{67.98} \\
LLaMA-2-7B-chat & 62.66 / \textbf{63.49} & 55.45 / \textbf{56.59} \\
LLaMA-3-8B-instruct & 67.42 / \textbf{68.03} & \textbf{45.29} / 44.39 \\ 
Avg. & 64.36 / 65.16 (\textbf{0.80↑}) & 56.10 / 56.32 (\textbf{0.22↑}) \\ \hline
\multicolumn{3}{c}{few-shot CoT} \\ \hline
Mistral-7B-instruct & 63.49 / \textbf{65.55} & 52.83 / \textbf{54.87} \\
LLaMA-2-7B-chat & 55.63 / \textbf{55.72} & 55.61 / \textbf{58.56} \\
LLaMA-3-8B-instruct & 66.38 / \textbf{68.47} & 67.24 / \textbf{68.47} \\
Avg. & 61.83 / 63.25 (\textbf{1.42↑}) & 58.56 / 60.63 (\textbf{2.07↑}) \\ \hline
\end{tabular}}
\caption{Results on the Commonsense Reasoning task for different models. \label{tab:commonsense_result}}
\vspace{-10pt}
\end{table}
\begin{table}[ht]
\centering
\resizebox{0.48\textwidth}{!}{
\begin{tabular}{lccc}
\hline
\multirow{2}{*}{Model} & \multicolumn{2}{c}{Symbolic Reasoning} \\ \cline{2-3} 
 & Coin Flip & Random Letter \\ \hline
\multicolumn{3}{c}{zero-shot CoT} \\ \hline
Mistral-7B-instruct & \textbf{50.80} / 48.95 & 18.33 / \textbf{19.66} \\
LLaMA-2-7B-chat & 52.20 / \textbf{55.75} & 17.33 / \textbf{17.66} \\
LLaMA-3-8B-instruct & 90.45 / \textbf{90.55} & 38.00 / \textbf{38.00} \\ 
Avg. & 64.48 / 65.10 (\textbf{0.62↑}) &  24.55 / 25.11 (\textbf{0.56↑}) \\ \hline
\multicolumn{3}{c}{few-shot CoT} \\ \hline
Mistral-7B-instruct & \textbf{78.73} / 78.19 & 33.67 / \textbf{33.67} \\
LLaMA-2-7B-chat & 50.74 / \textbf{50.84} & 29.67 / \textbf{30.00} \\
LLaMA-3-8B-instruct & 77.02 / \textbf{77.59} & 53.67 / \textbf{54.67} \\
Avg. & 68.83 / 68.87 (\textbf{0.04↑}) & 39.00 / 39.45 (\textbf{0.45↑}) \\ \hline
\end{tabular}}
\caption{Results on the Symbolic Reasoning task for different models. \label{tab:symbolic_result}}
\vspace{-10pt}
\end{table}

\subsection{Utility Evaluation}

In this section, we use the reading vector $v$ to control the generation of CoT reasoning paths. As shown in Table~\ref{tab:arithmetic_result},~\ref{tab:commonsense_result} and~\ref{tab:symbolic_result},
we present the performance comparison between our approach and baselines on each task. From these tables, we can see the following interesting findings.

\paragraph{(i) Our framework can guide/correct the reasoning direction of LLM.}
For example, for zero-shot CoT, our approach demonstrates superior performance, surpassing Mistral-7B-instruct by 4\% on the SVAMP dataset, and for few-shot CoT, our approach achieves a significant improvement over LLaMA-2-7B-chat by 2.95\% on the CSQA dataset. This is mainly because our approach demonstrates that our methodology effectively directs the correct reasoning path of LLM, thereby improving the accuracy of reasoning. This highlights the effectiveness our interpretability in CoT, which employs concept-level representation read operations to localize errors in CoT, followed by control operations to rectify LLMs’ reasoning paths.

\paragraph{(ii) Few-shot CoT have stereotypes in the reasoning phase.}

From these tables, we observe that, except for the commonsense reasoning task, our approach generally performs better on zero-shot control than con few-shot control. This is mainly due to the fact that it is affected by the few shot examples. 
To investigate why the performance of the few-shot concept is inferior to that of the zero-shot prompt, we analyzed instances where the few-shot CoT concept failed. As illustrated in figure ~\ref{fig:zero_few_control}, it shows that the model answered incorrectly in the red part by the few-shots learning while initially stating a correct fact of the question in the blue part. It was affected by the few shots that someone even flipped the coin. Afterward, this behavior reinforces the incorrect cognition by adding the few-shots control that the model is wrong initially stating a fact. After adding the zero-shot control, the model accurately answered the question by stating the correct facts. These results indicate that few-shot learning produces the stereotype when reasoning.

\liangeat{The results also show that the few-shots concept is inferior to the zero-shot concept in arithmetic Reasoning and symbolic reasoning, although few-shots perform well in commonsense reasoning. 
To investigate why the performance of the few-shot concept is inferior to that of the zero-shot prompt, we analyzed instances where the few-shot CoT concept failed, as illustrated in Figure ~\ref{fig:zero_few_control}. The figure shows that the model answered incorrectly in the red part by the few-shots learning while initially stating a correct fact of the question in the blue part. It was affected by the few shots that someone even flipped the coin. Afterwards, this behaviour reinforces the incorrect cognition by adding the few-shots control that the model is wrong initially stating a fact. After adding the zero-shot control, the model accurately answered the question by stating the correct facts. These results indicate that few-shot learning produces the stereotype when reasoning. }

\liangeat{
1) our approach can guide/correct the reasoning direction of LLM, thereby improving the accuracy of reasoning.
2) zero-shot CoT stimulated by prompt (\emph{``Let's think step by step''}) can activate the reasoning concept of LLM.
3) few-shot CoT, especially when only a few examples are given, simply activates the LLM concept of answer style.
}


\subsection{Interpretability Visualization}
To further demonstrate the effectiveness of our approach, we provide additional cases in three models on arithmetic, commonsense, and symbolic reasoning tasks in Appendix \ref{Appendix:case_demo} (See Figure \ref{fig:case_aqua_zero}, \ref{fig:case_strategyqa_few}, \ref{fig:few_shot_case_llama3_coin_flip}, and \ref{fig:coterrorexample}).

For zero-shot CoT, Figure~\ref{fig:case_aqua_zero} displays the real prediction of a case by the model both without control and control. 
From this figure, we can observe that the model is wrong when simplifying the equation, and the reasoning path is offset. As shown in the \textcolor{crimson}{red} part, the reason for the mistake of the model is that multiple steps have been performed in the process of simplification, which leads to calculation errors.
After adding control to the model, the direction of the reasoning path of the model will be corrected to achieve the calculation error that occurs when the control is not added.

For few-shots CoT, Figure~\ref{fig:case_strategyqa_few} demonstrates the real prediction of a case by the model both without control and with control on the strategyQA dataset. From this figure, we can observe that the model initially makes a correct inference regarding the elevation of Mount Fuji and the depth of the Sea of Japan. However, the model erroneously concludes that Mount Fuji would not be out of the sea due to a miscalculation in comparing the mountain’s height with the sea’s depth as illustrated in the \textcolor{crimson}{red} segment. After adding control to the model, the reasoning path leads to the correct conclusion that the top of Mount Fuji would indeed be out of the Sea of Japan, as highlighted in the \textcolor{seagreen}{green} segment. 

The results highlight our interpretability in CoT, which employs concept-level representation read operations to localize errors in CoT, followed by control operations to rectify LLMs’ reasoning paths. This intuitive interpretable analysis for CoT reasoning is helpful in tracing and controlling the errors LLMs made during the CoT process.

\liangeat{
\warn{FIXXIN: why are there two similar case here?}Additionally, Figure~\ref{fig:few_shot_case_llama3_coin_flip} also showcases that the model initially makes a correct inference step by step about the coin's state after each flip.  However, the error arises from not correctly accounting for the even number of flips by Kydin. After adding control, the model correctly reasons that the coin has flipped a total of two times, an even number, leading to the accurate conclusion that the coin remains heads up, as shown in the \textcolor{seagreen}{green} segment. These demonstrate the significance of control in ensuring the model follows the correct reasoning path.
}

\eat{
\subsection{Ablation Study}

\paragraph{Number of Few-shots.}

\paragraph{Quality of Reading Dataset.}
}

\section{Conclusion}
In this work, we propose a framework to analyze and understand the CoT reasoning from \emph{Hopfieldian view} under zero-shot and few-shot settings. Experimental results show our framework can improve the accuracy of reasoning, highlighting that our framework can provide intuitive and interpretable analysis for CoT reasoning, which is helpful in tracing and controlling the errors LLMs make during the CoT process, rectifying LLMs' reasoning paths, and enhancing CoT's transparency.

\section*{Limitations}
Our current problem formulation is primarily focused
on text data. We consider multi-modal scenarios, \emph{i.e.,} analyzing concepts from multiple different modalities as a future research direction.

\section*{Ethics Statement}
Our framework can provide intuitive and interpretable
analysis for CoT reasoning, which is helpful in tracing and controlling the errors LLMs made during the CoT process. This poses an impact on rectifying LLMs' reasoning paths and enhancing CoT's transparency. 

\liangeat{
Some of the limitations of our work are as follows:
\warn{1. Limited number of base models: The number of base models used was inadequate, preventing a thorough evaluation of model performance. 2. Lack of comparison across different shots: The study did not compare the performance of models under varying shots conditions, which restricts the comprehensiveness of the results.}
\warn{Fill details. Limitation is mandatory in ACL submissions.}
}

\bibliography{anthology,custom}
\bibliographystyle{acl_natbib}

\clearpage
\appendix
\section{Chain-of-Thought}
\label{cot}
\begin{figure}[ht]
    \centering
    \includegraphics[width=0.95\linewidth]{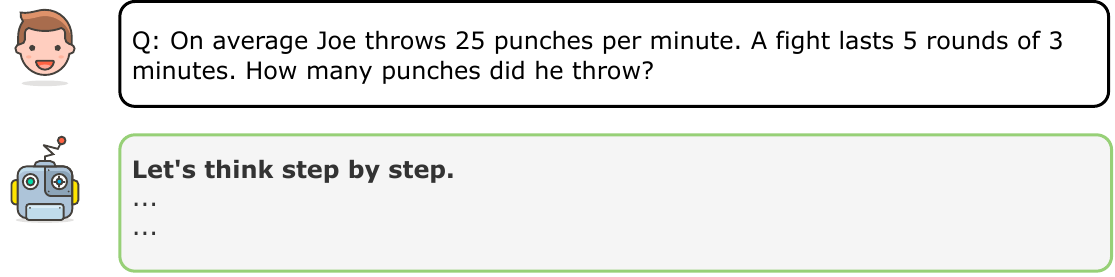}
    \caption{A example of using Zero-shot CoT.}
    \label{fig:app_zero_shot}
\vspace{-7pt}
\end{figure}

\paragraph{Zero-shot CoT.} Zero-shot Chain-of-Thought (CoT) \cite{kojima2022large} is a reasoning approach where a language model generates a step-by-step explanation or thought process to solve a problem without requiring prior examples or specific training. Given a problem $Q$, we aim to derive the answer using the zero-shot CoT method in a generate function $f$: 
\begin{equation}
\label{eq:function_1}
\vspace{-0.1ex}
A = f(Q, P).
\vspace{-0.1ex}
\end{equation}
where P = \emph{``Let' s think step by step''} in this research. The example of zero-shot CoT is shown in Figure~\ref{fig:app_zero_shot}

\paragraph{Few-shots CoT.} Few-shots Chain-of-Thought (CoT) \cite{wei2022chain} is a prompting technique used in large language models where a few examples of step-by-step reasoning processes are provided. Given a problem $Q$, we aim to derive the answer using the few-shot CoT method in a generate function $f$: 
\begin{equation}
\label{eq:function_2}
\vspace{-0.1ex}
A = f(T, Q).
\vspace{-0.1ex}
\end{equation}
where \( T = \{ t_1, t_2, \ldots, t_n \} \) and \( t_i \) represents the \( i \)-th example in the prompt. The example of zero-shot CoT shown on Figure~\ref{fig:app_few_shot}, \( T \) can be shown in Appendix~\ref{Appendix:Prompts_E}.
\begin{figure}[ht]
    \centering
    \includegraphics[width=0.95\linewidth]{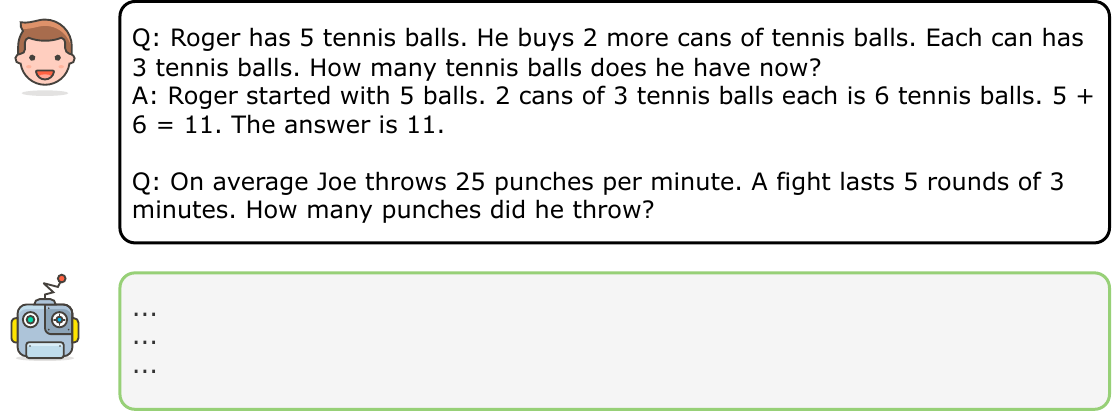}
    \caption{A example of using Few-shot CoT.}
    \label{fig:app_few_shot}
\vspace{-7pt}
\end{figure}

\section{Other Experimental Details}
\label{Appendix:exp}
\subsection{Dataset}
\label{Appendix:data}
The statistics of the data is shown in Table~\ref{tab:data_statistic}. The details about
each data set are as follows:

\paragraph{Arithmetic Reasoning.} The arithmetic reasoning 
benchmarks aim to analyze and/or understand the model's 
mathematical reasoning skills. These include: 
(i) GSM8K~\cite{cobbe2021gsm8k}, a math word problems benchmark 
encompassing a variety of reasoning steps;
(ii) SVAMP~\cite{patel-etal-2021-nlp}, containing math word 
problems with multiple structures;
(iii) AQuA~\cite{miao-etal-2020-diverse}, containing 
algebraic applications and mathematical reasoning problems.
\eat{The datasets above were used to assess the model's 
mathematical reasoning skills}

\paragraph{Commonsense Reasoning.} These data sets 
aim to analyze the ability of the model on commonsense 
reasoning tasks. These include:
(i) StrategyQA~\cite{10.1162/tacl_a_00370}, a commonsense benchmark 
requiring multi-level strategy to answer the question; 
\eat{(ii) Date Understanding~\cite{srivastava2023beyond}, a dataset 
that describes the ability of language models to understand the change in dates;}
(ii) CSQA~\cite{talmor2019commonsenseqa} benchmark dataset of multiple-choice questions that require different types of commonsense knowledge to predict the correct answers.

\liangeat{(iii) Causal Judgment~\cite{srivastava2023beyond}, a dataset that tests the model's ability to correctly determine causality.}

\eat{by describing 
a short story that introduces multiple causal events.}

\paragraph{Symbolic Reasoning.} These data sets aim to test 
the abilities of the model requiring advanced symbolic capabilities. 
For this task, we curated two new datasets, as follows.
(i) Coin Flip dataset, we employ the data curation strategy of a previous 
study~\cite{wei2022chain} using the number of 
operations as 2, 4 and 7 to come up with the complete dataset; 
(ii) Random Letter, an advanced version of the last letter concatenation with reference to the previously studied form of word assembly~\cite{wei2022chain}, 
where 2-4 words are randomly formed and characters are randomly drawn from them, instead of taking the beginning or the end of each word at a fixed point.

\begin{table}[ht]
\centering
\resizebox{\columnwidth}{!}{%
\begin{tabular}{cccc}
\hline
Dataset & Task Domain & $N$ & Answer Format \\ \hline
GSM8K & Arithmetic & 1319 & Number \\
SVAMP & Arithmetic & 300 & Number \\
AQuA & Arithmetic & 254 & Multiple Choices \\
StrategyQA & Commonsense & 2290 & Yes or No \\
CSQA & Commonsense & 1221 & Multiple Choices \\
Coin Flip & Symbolic & 2000 & Yes or No \\
Random Letter & Symbolic & 300 & Letter \\ \hline
\end{tabular}%
}
\caption{Statistics of the data set. $N$ represents the 
number of evaluation examples.}
\label{tab:data_statistic}
\end{table}

\subsection{The number of reading set sizes}
The table \ref{tab:ntrain_choices} below presents the sizes of reading sets utilized in our study providing the different set sizes employed during the reading phase on zero-shot and few-shots CoT.
\label{Appendix:reading_size}
\begin{table}[ht]
\centering
\resizebox{0.9\columnwidth}{!}{%
\begin{tabular}{cccc}
\hline
Dataset & Zero-shot & Few-shots  \\ \hline
GSM8K & 128 & 512 \\
SVAMP & 512 & 256  \\
AQuA & 256 & 256  \\
StrategyQA & 512 & 256 \\
CSQA & 512 & 256  \\
Coin Flip & 128 & 512 \\
Random Letter & 128 & 128 \\ \hline
\end{tabular}%
}
\caption{The number of reading set sizes in the experiment.}
\label{tab:ntrain_choices}
\end{table}

\section{Prompts}
\label{Appendix:prompts}

\subsection{Postive and Negative Stimulus}
\label{Appendix:pos_neg_s}
\begin{tcolorbox}[colback=white!15,colframe=pink!50!black,left=5pt,right=5pt,top=5pt,bottom=5pt]
\emph{positive stimulus} \\
USER: Would a greyhound be able to outrun a greyhound bus? \\
ASSISTANT: Let's think step by step. \\
\\
\emph{negative stimulus} \\
USER: Would a greyhound be able to outrun a greyhound bus? \\
ASSISTANT:
\end{tcolorbox}


\begin{table}[h]
\centering
\resizebox{\columnwidth}{!}{%
\begin{tabular}{cl}
\hline
\multirow{2}{*}{zero-shot CoT} & \begin{tabular}[c]{@{}l@{}}USER: <question>\\ ASSISTANT: \textcolor{red!70!black}{Let's think step by step.}\end{tabular} \\ \cline{2-2} 
 & \begin{tabular}[c]{@{}l@{}}USER: <question>\\ ASSISTANT:\end{tabular} \\ \hline
\multirow{2}{*}{few-shot CoT} & \begin{tabular}[c]{@{}l@{}}USER: \textcolor{red!70!black}{<$n$ different examples>}\\ <question>\\ ASSISTANT:\end{tabular} \\ \cline{2-2} 
 & \begin{tabular}[c]{@{}l@{}}USER: <question>\\ ASSISTANT:\end{tabular} \\ \hline
\end{tabular}%
}
\caption{The stimulus prompting design for CoT-style methods.}
\label{tab:prompt_for_stimulus}
\end{table}
\subsection{Prompt Templates}
\label{Appendix:Prompts_T}

Table~\ref{tab:prompt_for_stimulus} illustrates the design of stimulus prompts utilized for Chain of Thought (CoT) prompting, distinguishing between zero-shot CoT and few-shot CoT methodologies. In the zero-shot CoT approach, the model is presented with a question devoid of preceding examples, in contrast to the few-shot CoT method, where the model is furnished with multiple exemplars. For each method, the first row is a positive prompt and the second is a negative prompt. Red indicates stimulus token.


\begin{table}[ht]
\centering
\resizebox{\columnwidth}{!}{%
\begin{tabular}{cl}
\hline
\textbf{Task} & \textbf{Extraction Template} \\ \hline
GSM8K & Therefore, the answer (arabic numerals) is \\
SVAMP & Therefore, the answer (arabic numerals) is \\
AQuA & Therefore, among A through E, the answer is \\
StrategyQA & Therefore, the answer (Yes or No) is \\
CSQA & Therefore, among A through E, the answer is \\
Coin Flip & Therefore, the answer (Yes or No) is \\
Random Letter & Therefore, the answer is \\ \hline
\end{tabular}%
}
\caption{Extraction templates and answer cleansing approaches for various tasks.}
\label{tab:extraction_cleansing}
\end{table}
\begingroup
\begin{table*}[!ht]
    \centering
    \small
    \begin{tabular}{p{0.8\linewidth}}
        \toprule
        \underline{\textbf{\textsc{Prompt for GSM8K and SVAMP}}} \\
        \vspace{-2mm}
        \textbf{Q:} There are 15 trees in the grove. Grove workers will plant trees in the grove today. After they are done, there will be 21 trees. How many trees did the grove workers plant today? \\
        \vspace{-1mm}
        \textbf{A:} \hl{There are 15 trees originally. Then there were 21 trees after some more were planted. So there must have been 21 - 15 = 6. } The answer is 6. \\
        \vspace{0mm}
        \textbf{Q:} If there are 3 cars in the parking lot and 2 more cars arrive, how many cars are in the parking lot? \\
        \vspace{-1mm}
        \textbf{A:} \hl{There are originally 3 cars. 2 more cars arrive. 3 + 2 = 5.}  The answer is 5. \\
        \vspace{0mm}
        \textbf{Q:} Leah had 32 chocolates and her sister had 42. If they ate 35, how many pieces do they have left in total? \\
        \vspace{-1mm}
        \textbf{A:} \hl{Originally, Leah had 32 chocolates. Her sister had 42. So in total they had 32 + 42 = 74. After eating 35, they had 74 - 35 = 39.} The answer is 39. \\
        \vspace{0mm}
        \textbf{Q:} Jason had 20 lollipops. He gave Denny some lollipops. Now Jason has 12 lollipops. How many lollipops did Jason give to Denny? \\
        \vspace{-1mm}
        \textbf{A:} \hl{Jason started with 20 lollipops. Then he had 12 after giving some to Denny. So he gave Denny 20 - 12 = 8.} The answer is 8. \\
        \vspace{0mm}
        \textbf{Q:} Shawn has five toys. For Christmas, he got two toys each from his mom and dad. How many toys does he have now? \\
        \vspace{-1mm}
        \textbf{A:} \hl{Shawn started with 5 toys. If he got 2 toys each from his mom and dad, then that is 4 more toys. 5 + 4 = 9.} The answer is 9. \\
        \vspace{0mm}
        \textbf{Q:} There were nine computers in the server room. Five more computers were installed each day, from Monday to Thursday. How many computers are now in the server room? \\
        \vspace{-1mm}
        \textbf{A:} \hl{There were originally 9 computers. For each of 4 days, 5 more computers were added. So 5 * 4 = 20 computers were added. 9 + 20 is 29.} The answer is 29. \\
        \vspace{0mm}
        \textbf{Q:} Michael had 58 golf balls. On Tuesday, he lost 23 golf balls. On Wednesday, he lost 2 more. How many golf balls did he have at the end of Wednesday? \\
        \vspace{-1mm}
        \textbf{A:} \hl{Michael started with 58 golf balls. After losing 23 on Tuesday, he had 58 - 23 = 35. After losing 2 more, he had 35 - 2 = 33 golf balls.} The answer is 33. \\
        \vspace{0mm}
        \textbf{Q:} Olivia has \$23. She bought five bagels for \$3 each. How much money does she have left? \\
        \vspace{-1mm}
        \textbf{A:} \hl{Olivia had 23 dollars. 5 bagels for 3 dollars each will be 5 x 3 = 15 dollars. So she has 23 - 15 dollars left. 23 - 15 is 8.} The answer is 8. \\
        \bottomrule
    \end{tabular}
    \caption{Few-shot exemplars for full chain of thought prompt for the arithmetic reasoning task.}
    \label{tab:appendix-math-prompt}
\end{table*}
\endgroup

\subsection{Answer Extract Prompts}
\label{Appendix:extract}
The demonstration of our answer extraction method and answer clearing approach is in Table~\ref{tab:extraction_cleansing}.

\subsection{Example Prompts for Few-shot Setting}
\label{Appendix:Prompts_E}
We demonstrate our example prompts for few-shot setting in Table~\ref{tab:appendix-math-prompt}, \ref{tab:appendix-aqua-prompt}, \ref{tab:appendix-strategyqa-prompt}, \ref{tab:appendix-commonsenseqa-prompt}, \ref{tab:appendix-letter-concat-prompt} and \ref{tab:appendix-coinflip-prompt}.

\begingroup
\begin{table*}[!ht]
    \centering
    \small
    \begin{tabular}{p{0.8\linewidth}}
        \toprule
        \underline{\textbf{\textsc{Prompt for AQuA algebraic word problems}}} \\
        \vspace{-2mm}
        \textbf{Q:} John found that the average of 15 numbers is 40. If 10 is added to each number then the mean of the numbers is? \\ Answer Choices: (a) 50 (b) 45 (c) 65 (d) 78 (e) 64 \\
        \vspace{-1mm}
        \textbf{A:} \hl{If 10 is added to each number, then the mean of the numbers also increases by 10. So the new mean would be 50.} The answer is (a). \\
         \vspace{0mm}
        \textbf{Q:} If a / b = 3/4 and 8a + 5b = 22,then find the value of a. \\ Answer Choices: (a) 1/2 (b) 3/2 (c) 5/2 (d) 4/2 (e) 7/2 \\
        \vspace{-1mm}
        \textbf{A:} \hl{If a / b = 3/4, then b = 4a / 3. So 8a + 5(4a / 3) = 22. This simplifies to 8a + 20a / 3 = 22, which means 44a / 3 = 22. So a is equal to 3/2.} The answer is (b). \\
        \vspace{0mm}
        \textbf{Q:} A person is traveling at 20 km/hr and reached his destiny in 2.5 hr then find the distance? \\ Answer Choices: (a) 53 km (b) 55 km (c) 52 km (d) 60 km (e) 50 km \\
        \vspace{-1mm}
        \textbf{A:} \hl{The distance that the person traveled would have been 20 km/hr * 2.5 hrs = 50 km.}  The answer is (e). \\
        \vspace{0mm}
        \textbf{Q:} How many keystrokes are needed to type the numbers from 1 to 500? \\ Answer Choices: (a) 1156 (b) 1392 (c) 1480 (d) 1562 (e) 1788 \\
        \vspace{-1mm}
        \textbf{A:} \hl{There are 9 one-digit numbers from 1 to 9. There are 90 two-digit numbers from 10 to 99. There are 401 three-digit numbers from 100 to 500. 9 + 90(2) + 401(3) = 1392.} The answer is (b). \\
        \bottomrule
    \end{tabular}
    \caption{
    Few-shot exemplars for full chain of thought prompt for AQuA.
    }
    \label{tab:appendix-aqua-prompt}
\end{table*}
\endgroup
\begingroup
\begin{table*}[!ht]
    \centering
    \small
    \begin{tabular}{p{0.8\linewidth}}
        \toprule
        \underline{\textbf{\textsc{Prompt for StrategyQA}}} \\
        \vspace{-2mm}
        \textbf{Q:} Do hamsters provide food for any animals? \\
        \vspace{-1mm}
        \textbf{A:} \hl{Hamsters are prey animals. Prey are food for predators. Thus, hamsters provide food for some animals.} So the answer is yes. \\
        \vspace{0mm}
        \textbf{Q:} Could Brooke Shields succeed at University of Pennsylvania? \\
        \vspace{-1mm}
        \textbf{A:} \hl{Brooke Shields went to Princeton University. Princeton University is about as academically rigorous as the University of Pennsylvania. Thus, Brooke Shields could also succeed at the University of Pennsylvania.}  So the answer is yes. \\
        \vspace{0mm}
        \textbf{Q:} Yes or no: Hydrogen's atomic number squared exceeds number of Spice Girls? \\
        \vspace{-1mm}
        \textbf{A:} \hl{Hydrogen has an atomic number of 1. 1 squared is 1. There are 5 Spice Girls. Thus, Hydrogen's atomic number squared is less than 5.} So the answer is no. \\
        \vspace{0mm}
        \textbf{Q:} Yes or no: Is it common to see frost during some college commencements? \\
        \vspace{-1mm}
        \textbf{A:} \hl{College commencement ceremonies can happen in December, May, and June. December is in the winter, so there can be frost. Thus, there could be frost at some commencements.} So the answer is yes. \\
        \vspace{0mm}
        \textbf{Q:} Yes or no: Could a llama birth twice during War in Vietnam (1945-46)? \\
        \vspace{-1mm}
        \textbf{A:} \hl{The War in Vietnam was 6 months. The gestation period for a llama is 11 months, which is more than 6 months. Thus, a llama could not give birth twice during the War in Vietnam.} So the answer is no. \\
        \vspace{0mm}
        \textbf{Q:} Yes or no: Would a pear sink in water? \\
        \vspace{-1mm}
        \textbf{A:} \hl{The density of a pear is about $0.6 g/cm^3$, which is less than water. Objects less dense than water float. Thus, a pear would float.} So the answer is no. \\
        \bottomrule
    \end{tabular}
    \caption{Few-shot exemplars for full chain of thought prompt for StrategyQA.}
    \label{tab:appendix-strategyqa-prompt}
\end{table*}
\endgroup
\begingroup
\begin{table*}[!ht]
    \centering
    \small
    \begin{tabular}{p{0.8\linewidth}}
        \toprule
        \underline{\textbf{\textsc{Prompt for CSQA}}} \\
        \vspace{-2mm}
        \textbf{Q:} What do people use to absorb extra ink from a fountain pen? Answer Choices: (a) shirt pocket (b) calligrapher's hand (c) inkwell (d) desk drawer (e) blotter \\
        \vspace{-1mm}
        \textbf{A:} \hl{The answer must be an item that can absorb ink. Of the above choices, only blotters are used to absorb ink.} So the answer is (e). \\
        \vspace{0mm}
        \textbf{Q:} What home entertainment equipment requires cable? \\
        Answer Choices: (a) radio shack (b) substation (c) television (d) cabinet \\
        \vspace{-1mm}
        \textbf{A:} \hl{The answer must require cable. Of the above choices, only television requires cable.} So the answer is (c). \\
        \vspace{0mm}
        \textbf{Q:} The fox walked from the city into the forest, what was it looking for? Answer Choices: (a) pretty flowers (b) hen house (c) natural habitat (d) storybook\\
        \vspace{-1mm}
        \textbf{A:} \hl{The answer must be something in the forest. Of the above choices, only natural habitat is in the forest.}  So the answer is (b). \\
        \vspace{0mm}
        \textbf{Q:} Sammy wanted to go to where the people were. Where might he go? Answer Choices: (a) populated areas (b) race track (c) desert (d) apartment (e) roadblock \\
        \vspace{-1mm}
        \textbf{A:} \hl{The answer must be a place with a lot of people. Of the above choices, only populated areas have a lot of people.} So the answer is (a). \\
        \vspace{0mm}
        \textbf{Q:} Where do you put your grapes just before checking out? Answer Choices: (a) mouth (b) grocery cart (c)super market (d) fruit basket (e) fruit market\\
        \vspace{-1mm}
        \textbf{A:} \hl{The answer should be the place where grocery items are placed before checking out. Of the above choices, grocery cart makes the most sense for holding grocery items.} So the answer is (b). \\
        \vspace{0mm}
        \textbf{Q:} Google Maps and other highway and street GPS services have replaced what? Answer Choices: (a) united states (b) mexico (c) countryside (d) atlas \\
        \vspace{-1mm}
        \textbf{A:} \hl{The answer must be something that used to do what Google Maps and GPS services do, which is to give directions. Of the above choices, only atlases are used to give directions.} So the answer is (d). \\
        \vspace{0mm}
        \textbf{Q:} Before getting a divorce, what did the wife feel who was doing all the work? Answer Choices: (a) harder (b) anguish (c) bitterness (d) tears (e) sadness \\
        \vspace{-1mm}
        \textbf{A:} \hl{The answer should be the feeling of someone getting divorced who was doing all the work. Of the above choices, the closest feeling is bitterness.} So the answer is (c). \\
        \bottomrule
    \end{tabular}
    \caption{Few-shot exemplars for full chain of thought prompt for CSQA. There are newlines between the answer choices that are omitted in the table for space reasons.}
    \label{tab:appendix-commonsenseqa-prompt}
\end{table*}
\endgroup
\begingroup
\begin{table*}[!ht]
    \centering
    \small
    \begin{tabular}{p{0.8\linewidth}}
        \toprule
        \underline{\textbf{\textsc{Prompt for Rondom Letter Concatenation}}} \\
        \vspace{-2mm}
        \textbf{Q:} Take the first letters of the words in "James Hickman
" and concatenate them. \\
        \vspace{-1mm}
        \textbf{A:} \hl{The first letter of "James" is "J". The first letter of "Hickman" is "H". Concatenating them is "JH".} The answer is JH. \\
         \vspace{0mm}
        \textbf{Q:} Take the second
 letters of the words in "Carl Mccall Bonilla
" and concatenate them. \\
        \vspace{-1mm}
        \textbf{A:} \hl{The second letter of "Carl" is "a". The second letter of "Mccall" is "c". The second letter of "Bonilla" is "o" Concatenating them is "aco".} The answer is aco. \\
        \vspace{0mm}
        \textbf{Q:} Take the third letters of the words in "Randy Tanner
" and concatenate them. \\
        \vspace{-1mm}
        \textbf{A:} \hl{The third letter of "Randy" is "n". The third letter of "Tanner" is "n". Concatenating them is "nn".}  The answer is nn. \\
        \vspace{0mm}
        \textbf{Q:} Take the first letters of the words in "Kenny Kim
" and concatenate them. \\
        \vspace{-1mm}
        \textbf{A:} \hl{The first letter of "Kenny" is "K". The first letter of "Kim" is "K". Concatenating them is "KK".} The answer is KK. \\
        \bottomrule
    \end{tabular}
    \caption{Few-shot exemplars for full chain of thought prompt for the random letter concatenation task.}
    \label{tab:appendix-letter-concat-prompt}
\end{table*}
\endgroup
\begingroup
\begin{table*}[!ht]
    \centering
    \small
    \begin{tabular}{p{0.8\linewidth}}
        \toprule
        \underline{\textbf{\textsc{Prompt for Coin Flip}}} \\
        \vspace{-2mm}
        \textbf{Q:} A coin is heads up. Ka flips the coin. Sherrie flips the coin. Is the coin still heads up? \\
        \vspace{-1mm}
        \textbf{A:} \hl{The coin was flipped by Ka and Sherrie. So the coin was flipped 2 times, which is an even number. The coin started heads up, so after an even number of flips, it will still be heads up.} So the answer is yes. \\
        \vspace{0mm}
        \textbf{Q:} A coin is heads up. Jamey flips the coin. Teressa flips the coin. Is the coin still heads up? \\
        \vspace{-1mm}
        \textbf{A:} \hl{The coin was flipped by Jamey and Teressa. So the coin was flipped 2 times, which is an even number. The coin started heads up, so after an even number of flips, it will still be heads up.} So the answer is yes. \\
        \vspace{0mm}
        \textbf{Q:} A coin is heads up. Maybelle flips the coin. Shalonda does not flip the coin. Is the coin still heads up? \\
        \vspace{-1mm}
        \textbf{A:} \hl{The coin was flipped by Maybelle. So the coin was flipped 1 time, which is an odd number. The coin started heads up, so after an odd number of flips, it will be tails up.} So the answer is no. \\
        \vspace{0mm}
        \textbf{Q:} A coin is heads up. Millicent does not flip the coin. Conception flips the coin. Is the coin still heads up? \\
        \vspace{-1mm}
        \textbf{A:} \hl{The coin was flipped by Conception. So the coin was flipped 1 time, which is an odd number. The coin started heads up, so after an odd number of flips, it will be tails up.} So the answer is no. \\
        \vspace{0mm}
        \textbf{Q:} A coin is heads up. Sal flips the coin. Raymond does not flip the coin. Is the coin still heads up? \\
        \vspace{-1mm}
        \textbf{A:} \hl{The coin was flipped by Sal. So the coin was flipped 1 time, which is an odd number. The coin started heads up, so after an odd number of flips, it will be tails up.} So the answer is no. \\
        \vspace{0mm}
        \textbf{Q:} A coin is heads up. Conception flips the coin. Kristian does not flip the coin. Is the coin still heads up? \\
        \vspace{-1mm}
        \textbf{A:} \hl{The coin was flipped by Conception. So the coin was flipped 1 time, which is an odd number. The coin started heads up, so after an odd number of flips, it will be tails up.} So the answer is no. \\
        \vspace{0mm}
        \textbf{Q:} A coin is heads up. Inga does not flip the coin. Elanor does not flip the coin. Is the coin still heads up? \\
        \vspace{-1mm}
        \textbf{A:} \hl{The coin was flipped by no one. So the coin was flipped 0 times. The coin started heads up, and it was not flipped, so it is still heads up.} So the answer is yes. \\
        \vspace{0mm}
        \textbf{Q:} A coin is heads up. Ryan flips the coin. Shaunda flips the coin. Is the coin still heads up? \\
        \vspace{-1mm}
        \textbf{A:} \hl{The coin was flipped by Ryan and Shaunda. So the coin was flipped 2 times, which is an even number. The coin started heads up, so after an even number of flips, it will still be heads up.} So the answer is yes. \\
        \bottomrule
    \end{tabular}
    \caption{Few-shot exemplars for full chain of thought prompt for the coin flip task.}
    \label{tab:appendix-coinflip-prompt}
\end{table*}
\endgroup

\newpage
\section{Case Demonstrations}
\label{Appendix:case_demo}

\begin{figure*}[t]
    \centering
    \includegraphics[width=0.9\linewidth]{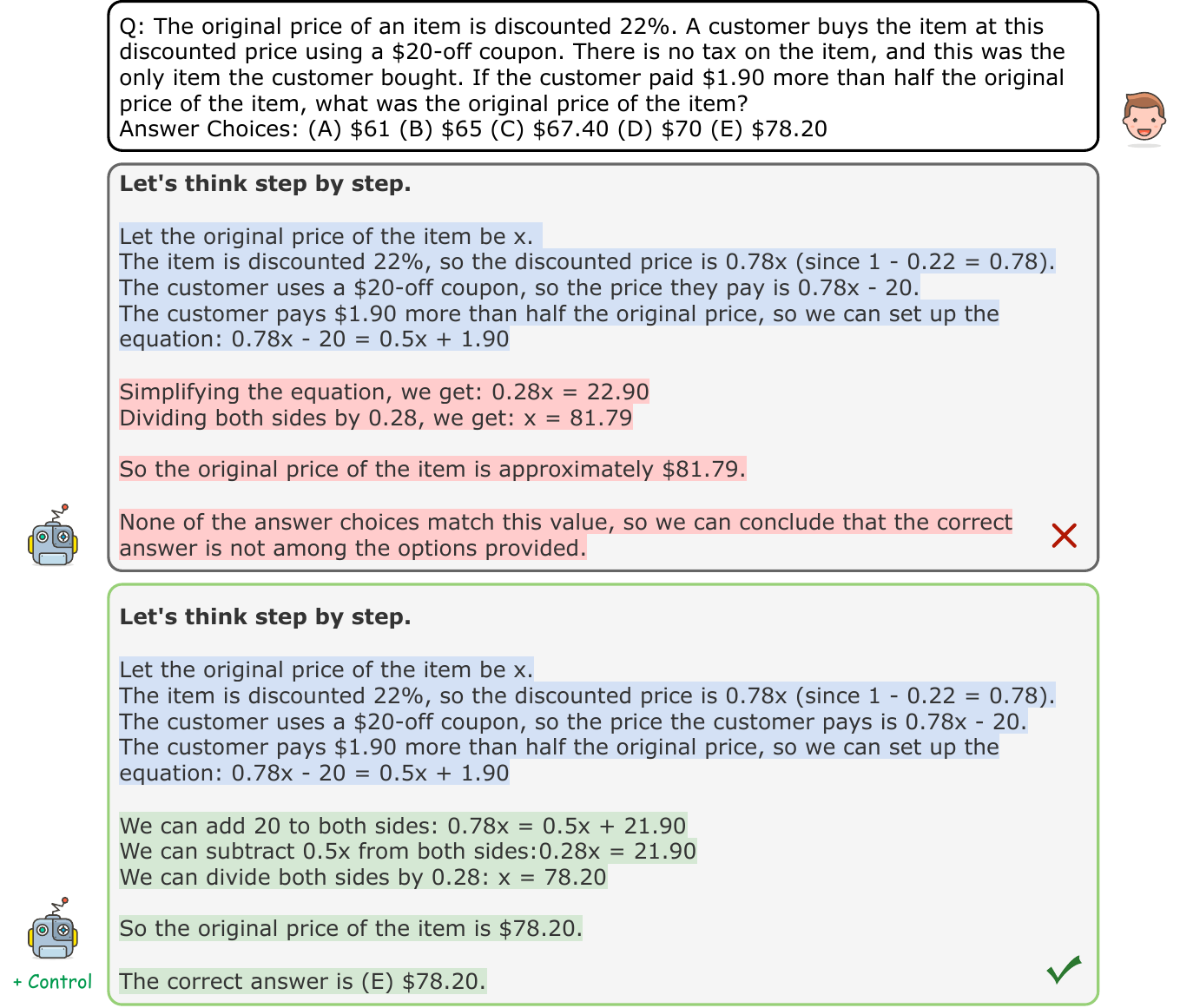}
    \caption{A real case predicted by LLaMA-3-8B-instruct with zero-shot CoT on the AQuA dataset. 
    The segment highlighted in \textcolor{skyblue}{blue} represents the correct output of the model. 
    The \textcolor{crimson}{red} part shows that the model starts to reason in the wrong direction without control, 
    while the \textcolor{seagreen}{green} portion indicates the model reason in the correct direction after adding control.}
    \label{fig:case_aqua_zero}
\end{figure*}

\begin{figure*}[ht]
    \centering
    \includegraphics[width=0.9\linewidth]{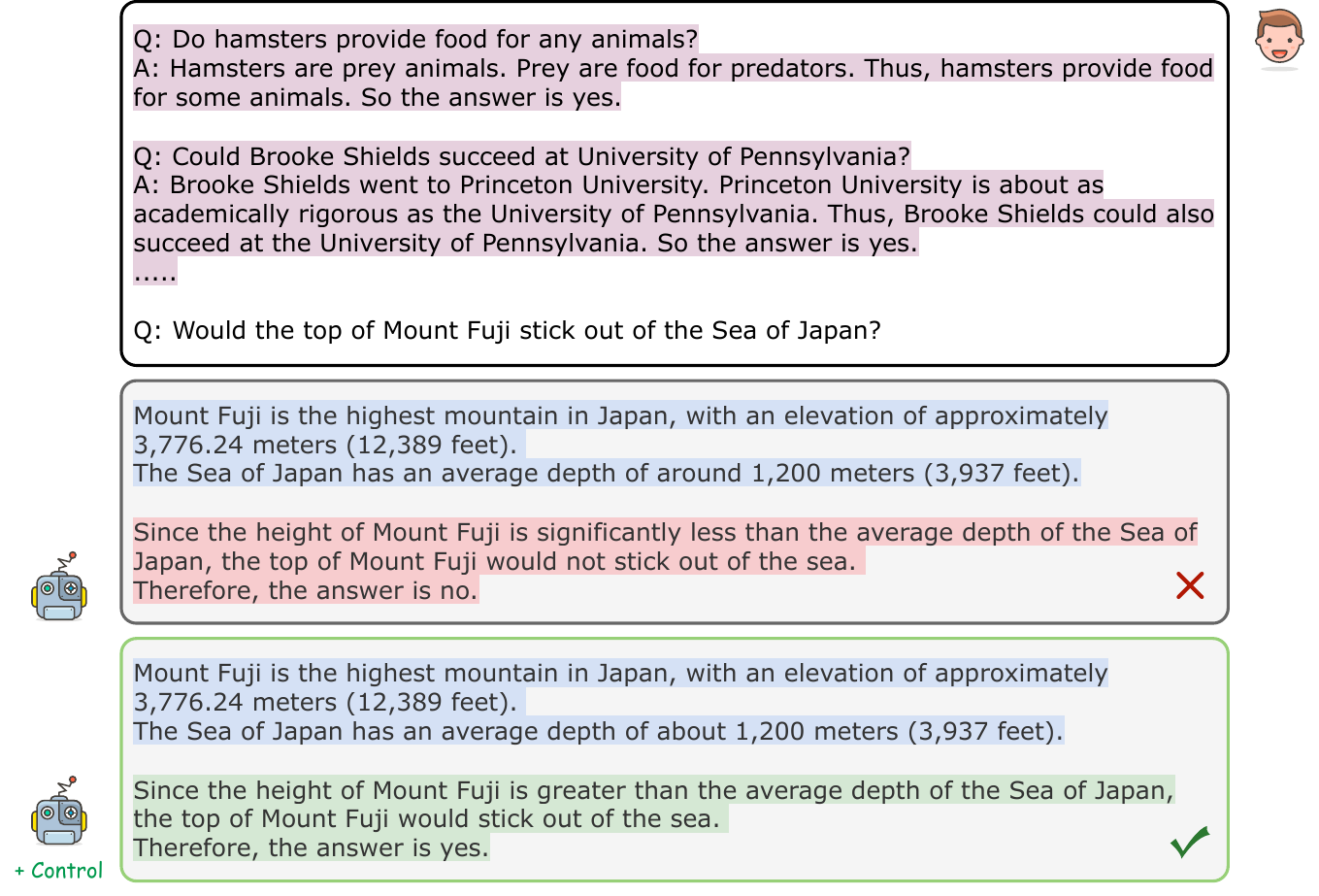}
    \caption{A real case predicted by Mistral-7B-Instruct-v0.2 with few-shots CoT on the strategyQA dataset. 
    The \textcolor{purple}{purple} part is an example of input-output pairs given by user.
    The segment highlighted in \textcolor{skyblue}{blue} represents the correct output of the model. 
    The \textcolor{crimson}{red} part shows that the model starts to reason in the wrong direction without control, 
    while the \textcolor{seagreen}{green} portion indicates the model reason in the correct direction after adding control.}
    \label{fig:case_strategyqa_few}
\end{figure*}

\begin{figure*}[ht]
    \centering
    \includegraphics[width=0.9\linewidth]{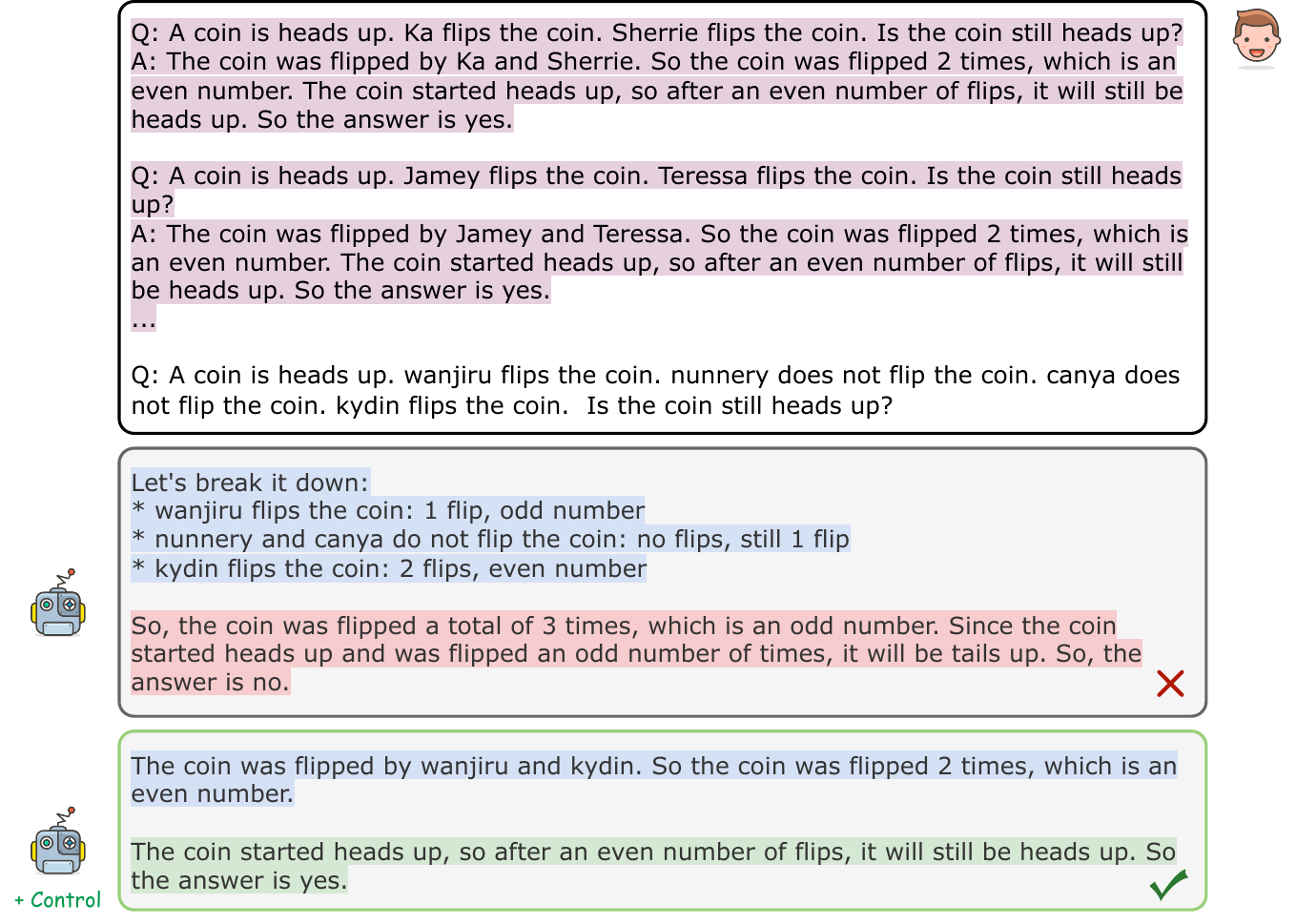}
    \caption{A real case predicted by Meta-Llama-3-8B-Instruct with few-shots CoT on the coin flip dataset. 
    The \textcolor{purple}{purple} part is an example of input-output pairs given by user.
    The segment highlighted in \textcolor{skyblue}{blue} represents the correct output of the model. 
    The \textcolor{crimson}{red} part shows that the model starts to reason in the wrong direction without control, 
    while the \textcolor{seagreen}{green} portion indicates the model reason in the correct direction after adding control.}
    \label{fig:few_shot_case_llama3_coin_flip}
\end{figure*}

\begin{figure*}
    \centering
    \includegraphics[width=0.90\textwidth]{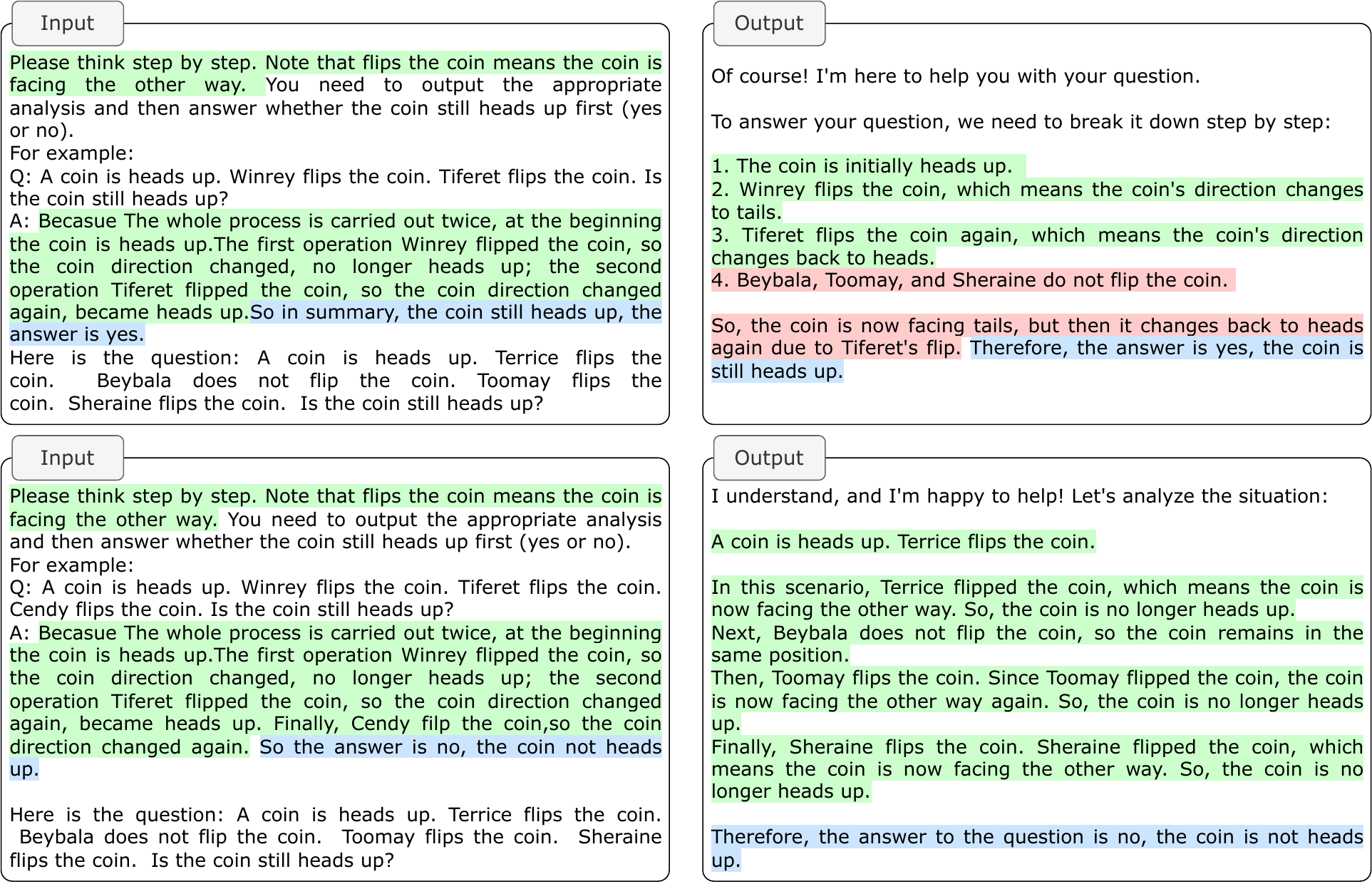}
    \caption{
    An example illustrating the sensitivity of LLaMA-2-7B-chat
    in the one-shot scenario. 
    The parts highlighted in \textcolor{springgreen}{\textbf{green}} are the instruction and reasoning path.
    The key difference between the two input-output pairs is highlighted in \textcolor{skyblue}{\textbf{blue}}: The example in the first row has a final answer of ``yes'', while the example in the second row has a final answer of ``no''. We can find that the final output of the model tends to \textit{match} the example, even if the example's result is incorrect. The model adjusts its reasoning steps (highlighted in \textcolor{salmon}{\textbf{red}}) to achieve this consistency.} 
    \vspace{-3.1ex}
    \label{fig:coterrorexample}
\end{figure*}

We show our additional cases in three models on arithmetic, commonsense, and symbolic reasoning tasks in Figure \ref{fig:case_aqua_zero}, \ref{fig:case_strategyqa_few}, \ref{fig:few_shot_case_llama3_coin_flip}, and \ref{fig:coterrorexample}.


\end{document}